\DocumentMetadata{}
\documentclass[sigconf]{acmart}

\usepackage{listings}
\usepackage{colortbl}
\usepackage{algorithm}
\usepackage{algpseudocode}
\usepackage{hyperxmp}

\definecolor{codegreen}{rgb}{0,0.6,0}
\definecolor{codegray}{rgb}{0.5,0.5,0.5}
\definecolor{codepurple}{rgb}{0.58,0,0.82}
\definecolor{backcolour}{rgb}{0.95,0.95,0.92}

\newcommand{\etal}{et al.\ }

\lstdefinestyle{mystyle}{
    commentstyle=\color{codegreen},
    keywordstyle=\color{magenta},
    numberstyle=\tiny\color{codegray},
    stringstyle=\color{codepurple},
    basicstyle=\ttfamily\footnotesize,
    breakatwhitespace=false,         
    breaklines=true,                 
    captionpos=b,                    
    keepspaces=true,                 
    numbers=left,                    
    numbersep=5pt,                  
    showspaces=false,                
    showstringspaces=false,
    showtabs=false,                  
    tabsize=2
}

\AtBeginDocument{%
  \providecommand\BibTeX{{%
    \normalfont B\kern-0.5em{\scshape i\kern-0.25em b}\kern-0.8em\TeX}}}

\copyrightyear{2026}
\acmYear{2026}
\acmConference[UIST '26]{The 39th Annual ACM Symposium on User Interface Software and Technology}{November 02--05, 2026}{Detroit, MI, USA}
\acmBooktitle{The 39th Annual ACM Symposium on User Interface Software and Technology (UIST '26), November 02--05, 2026, Detroit, MI, USA}
\acmDOI{10.1145/3830398.3830647}
\acmISBN{979-8-4007-2856-3/2026/11}

\usepackage{soul}

\usepackage{booktabs}
\usepackage{multirow}
\usepackage{graphicx}
\usepackage{microtype}

  \usepackage{xcolor}

\def\changesarevisible{0}

\ifx\changesarevisible\undefined
undefined
\else
  \if\changesarevisible1
    \newcommand{\finalAdd}[1]{\textcolor{blue}{#1}}
    \newcommand{\finalDel}[1]{\textcolor{magenta}{\st{#1}}}
    
    \newcommand{\finalReplace}[2]{\finalDel{#1} \finalAdd{#2}}
    
  \else
    \newcommand{\finalAdd}[1]{#1}
    \newcommand{\finalDel}[1]{}
    
    \newcommand{\finalReplace}[2]{\finalAdd{#2}}
    
  \fi
\fi

\begin{document}




\title[InstructMesh]{InstructMesh: Selective Refinement of Generative 3D Models for Fabrication}


\author{Faraz Faruqi}
\email{ffaruqi@mit.edu}
\orcid{0000-0002-1691-2093}
\affiliation{%
  \institution{MIT CSAIL}
  \city{Cambridge}
  \state{MA}
  \country{USA}
}

\author{Ahmed Katary}
\email{atkatary@google.com}
\affiliation{%
  \institution{Google}
  \city{Sunnyvale}
  \state{CA}
  \country{USA}
}

\author{Demircan Tas}
\email{tasd@mit.edu}
\affiliation{%
  \institution{MIT CSAIL}
  \city{Cambridge}
  \state{MA}
  \country{USA}
}

\author{Theresa Hradilak}
\email{hradilak@mit.edu}
\affiliation{%
  \institution{MIT CSAIL}
  \city{Cambridge}
  \state{MA}
  \country{USA}
}

\author{Ning Zhang}
\email{ningrz@mit.edu}
\affiliation{%
  \institution{MIT CSAIL}
  \city{Cambridge}
  \state{MA}
  \country{USA}
}

\author{Jiaji Li}
\email{jiaji@mit.edu}
\affiliation{%
  \institution{MIT CSAIL}
  \city{Cambridge}
  \state{MA}
  \country{USA}
}

\author{Fabian Manhardt}
\email{fabianmanhardt@google.com}
\affiliation{%
  \institution{Google}
  \city{Munich}
  \country{Germany}
}

\author{Martin Nisser}
\email{nisser@uw.edu}
\affiliation{%
  \institution{University of Washington}
  \city{Seattle}
  \state{WA}
  \country{USA}
}

\author{Vrushank Phadnis}
\email{vrushank@google.com}
\affiliation{%
  \institution{Google}
  \city{Mountain View}
  \state{CA}
  \country{USA}
}

\author{Ruofei Du}
\email{ruofei@google.com}
\affiliation{%
  \institution{Google XR}
  \city{San Francisco}
  \state{CA}
  \country{USA}
}

\author{Federico Tombari}
\email{tombari@google.com}
\affiliation{%
  \institution{Google}
  \city{Zurich}
  \country{Switzerland}
}

\author{Megan Hofmann}
\email{m.hofmann@northeastern.edu}
\orcid{0000-0003-2283-8587}
\affiliation{%
  \institution{Khoury College of Computer Sciences, Northeastern University}
  \city{Boston}
  \state{MA}
  \country{USA}
}

\author{Stefanie Mueller}
\email{stefmue@mit.edu}
\orcid{0000-0001-7743-7807}
\affiliation{%
  \institution{MIT CSAIL}
  \city{Cambridge}
  \state{MA}
  \country{USA}
}

\renewcommand{\shortauthors}{Faruqi et al.}



\begin{abstract}

Recent advances in generative AI allow users to create 3D models from text or images. However, these models prioritize visual plausibility over geometric accuracy, often generating results with flaws that compromise their intended use post-fabrication. We present InstructMesh, \finalReplace{a system}{an interactive post-generation refinement tool} that enables selective \finalReplace{refinement}{repair} of generative 3D models through region selection and targeted operations, such as opening or sealing voids, or adjusting local thickness. Users can invoke edit operations via natural language prompts or slider controls. By operating directly on the intermediate latent representation, InstructMesh allows users to apply robust geometric corrections without requiring expert modeling skills. To inform our design, we first analyze common fabrication-related failure modes in outputs from state-of-the-art generative tools. We then conduct two user studies\finalDel{with novices}, demonstrating that \finalAdd{novices can identify and perform fabrication-relevant repairs on generative outputs using} InstructMesh, \finalDel{improves fabrication viability of generative outputs}and revealing user preference for hybrid interfaces that combine slider controls with natural language input.

 \end{abstract}



\begin{teaserfigure}
\centering
  \includegraphics[width=\textwidth]{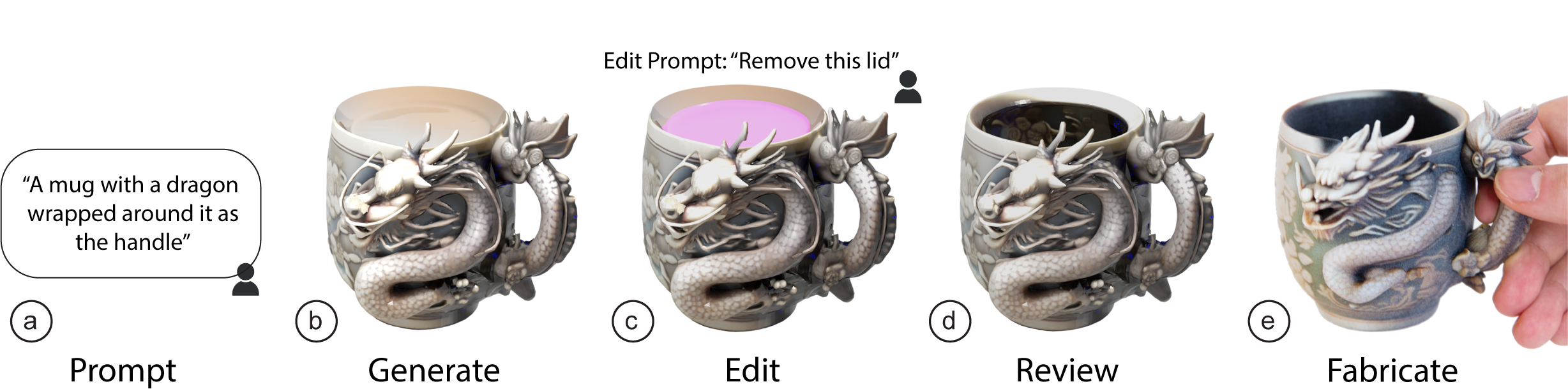}
  \vspace{-12pt}
  \caption{InstructMesh enables fabrication-relevant refinement of generative 3D models through selective editing. a)~A user provides an initial prompt. b)~The system generates a 3D model, but may include fabrication-related flaws (e.g., a sealed lid). c)~The user selects the problematic region and provides a descriptive edit prompt. d)~InstructMesh applies an edit operation in the latent-space to remove the lid while preserving other regions and regenerates the model. e)~The final model is fabricated as a functional mug.}
  \label{fig:teaser}
\end{teaserfigure}

\maketitle

\section{Introduction}

Generative AI has enabled new forms of creative expression across modalities like text~\cite{chowdhery2023palm}, images~\cite{rombach2022high}, music~\cite{AI-music}, and 3D geometry~\cite{gao2022get3d}. In the 3D domain, these advances now allow users with minimal modeling experience to generate novel 3D models from text or image prompts, bypassing the steep learning curve of traditional 3D modeling tools. However, while generative models produce visually plausible outputs, they are typically trained with image-based supervision, prioritizing appearance over the geometric properties required for real-world fabrication~\cite{faruqi2023style2fab}. As a result, generated models could exhibit flaws such as sealed openings, insufficient wall thickness, or missing structural connections, that only become apparent when a model is fabricated and used in a downstream physical task.
 
Bridging the gap between a visually plausible model and a \finalReplace{fabrication-ready}{physically usable} one requires \finalReplace{geometric corrections}{fabrication-relevant repairs} that are difficult for novices to achieve. Mesh-editing tools such as Blender~\cite{blenderUI} or MeshMixer~\cite{schmidt2010meshmixer} can address geometric flaws, but require significant domain expertise and effort that novices may not have. Recent text-driven 3D editing methods such as DreamEditor~\cite{dreameditor2023} and TIP-Editor~\cite{tipeditor2024} allow localized edits guided by text prompts, but operate on neural radiance fields to target visual appearance rather than the fabrication-specific geometric corrections \finalReplace{that are needed to make a model physically usable}{needed for real-world use}. Similarly, tools like Style2Fab~\cite{faruqi2023style2fab} and TactStyle~\cite{faruqi2025tactstyle} support functionality-aware transformations of existing designs, but assume that the relevant functional geometry already exists in the base model, focusing on preservation rather than repair. Finally, re-prompting the generative model is often unreliable: prior work shows that non-experts struggle to craft effective prompts~\cite{zamfirescu2023johnny,subramonyam2023bridging}, often producing vague or overly rigid instructions that fix one issue, only to alter other desired details, trapping them in a frustrating prompt-refinement loop~\cite{liu20233dall}. 
 
To bridge this gap, we introduce InstructMesh, \finalReplace{a system}{an interactive post-generation refinement tool} that \finalReplace{enables fabrication-relevant refinement}{supports fabrication-relevant repairs} of generative 3D models by mapping user-specified edits to targeted operations on the model's intermediate latent representation. Rather than forcing novices to edit the mesh directly, InstructMesh modifies the coarse voxel grid produced by the generative model's encoder, then passes the edited representation through the decoder to produce a refined mesh. This allows the generative model to fill in low-level details such as texture and surface topology, while satisfying the user's intended geometric correction.
 
A key challenge in designing such tools for generative 3D models is determining how novices should express geometric corrections: through natural language, which is expressive but ambiguous, or through structured controls, which are precise but require understanding the available operations? To investigate this, InstructMesh provides two complementary interaction modes: a natural language interface through an LLM, and a slider interface exposing a library of parameterized operations. In both modes, users first highlight a problematic region on the mesh, then specify the desired change. A second design challenge arises from the opacity of latent-space editing: because the intermediate representation is non-visual, users cannot anticipate how their edit will affect the final mesh. To make this process controllable, InstructMesh provides a preview visualization — showing additive edits in green and subtractive edits in red, making the latent-space edit tangible before regeneration is triggered. This introduces WYSIWYG\footnote{What You See Is What You Get}-style interaction into generative 3D modeling, allowing users to inspect, revise, and confirm edits on the intermediate representation, thereby maintaining direct control over the generative process. Together, these design choices enable us to investigate two research questions about novice-driven repair of generative 3D models for fabrication. First, can novices reliably identify and correct \finalAdd{visually identifiable} fabrication-relevant flaws using latent-space editing, a task that has traditionally required sophisticated mesh-manipulation skills? Second, how do the choice of interaction mode and the availability of visual feedback on the intermediate representation shape their editing experience?

In summary, this work contributes:
\begin{enumerate}
    \item A formative analysis (N = 120 models) identifying common fabrication-relevant geometric flaws in outputs from state-of-the-art 3D generative models,
    \item A set of canonical latent-space operations that repair these flaws without requiring mesh manipulation expertise, paired with an interaction design combining natural language input and parameterized slider controls, and a preview visualization for user understanding,
    \item Two user studies (N=12) demonstrating that novices can successfully identify and repair \finalAdd{visually identifiable} flaws using InstructMesh, and that hybrid workflows with preview visualizations improve usability and transparency.
\end{enumerate}

\section{Related Work}

We situate our work at the intersection of human-computer interaction, generative AI, and computational fabrication. Specifically, we draw on prior research in three key areas: (1) support tools in digital fabrication workflows, (2) 3D generative AI workflows and their integration into design processes, and (3) human-in-the-loop customization techniques in 2D and 3D generative systems.

%

\subsection{Supporting Users in Fabrication Workflows}

3D printing is emerging as makers' digital fabrication tool of choice~\cite{Kuznetsov_2010_expertamateur}. Online repositories such as Thingiverse\footnote{Thingiverse: \url{https://www.thingiverse.com}} provide a low barrier to entry for exploring 3D printing as users can quickly download pre-made 3D designs and explore design-specific questions~\cite{faruqi2021slicehub}. However, numerous studies have explored the practices of these novice makers~\cite{hudson2016understanding, norouzi2021making, berman2021howdiy, schmidt2013design} and tend to find that makers struggle with modifications to existing designs due to limitations of current computer-aided design (CAD) workflows. Oehlberg \etal~\cite{oehlberg2015patterns} observed that, even when designs are customizable, they lack the full range of modifiable features makers seek. 

Researchers have proposed several support tools that enable editing of 3D models by users with the goal to fabricate an object. Meshmixer~\cite{schmidt2010meshmixer} follows the approach of `Design to Fabricate'~\cite{schmidt2013design} and allows users to manipulate 3D models with the intent to 3D print them afterwards. Several parametric design tools~\cite{shugrina2015fab, schulz2014design, veuskens2020coda, stemasov2024param} allow interactive customization of parametric designs by a novice user, while maintaining validity and manufacturability of the design. Tools also support specific retargeting of mechanical components~\cite{zhang2017functionality, roumen2018grafter, koyama2015autoconnect}, allowing users to recombine mechanical components between 3D models. Several optimization algorithms target post-fabrication physical properties of 3D models, such as balance~\cite{prevost2013make}, moment-of-inertia for spinnable models~\cite{bacher2014spin}, or constrained optimization between center-of-mass and balance~\cite{zhao2016make, prevost2016balancing}. \finalAdd{Recent work has also demonstrated augmentation of generative models with mechanical simulation to ensure structural viability~\cite{faruqi2025mechstyle}}. 

However, these fabrication support tools typically edit existing meshes or models, limiting the ability of novice users to freely explore new designs. Recent developments in generative AI workflows offer an alternative approach, enabling users to construct bespoke 3D models directly from image and text prompts.


\subsection{3D Generative AI Workflows}

Reconstructing 3D objects from a single image or text description is a longstanding challenge in computer vision. Recent machine learning methods address this problem by learning priors over large collections of 3D shapes to infer the full geometry of an object from a single image. Several approaches have been proposed, including GANs~\cite{cai2020learning, gao2022get3d}, flow networks~\cite{klokov2020discrete, yang2019pointflow}, VAEs~\cite{mittal2022autosdf, wu2019sagnet}, and diffusion models~\cite{chou2023diffusion, jun2023shap_e}. The recent Large-Reconstruction Model (LRM) architecture~\cite{hong2023lrm} and its variants~\cite{xu2024instantmesh, tang2024lgm, boss2024sf3d} have demonstrated generalizable, high-quality 3D reconstruction. Among these, Trellis~\cite{xiang2024structured3dlatentsscalable} introduces a structured latent representation that separates geometry encoding from mesh decoding, achieving state-of-the-art reconstruction fidelity. Commercial platforms such as Meshy~\cite{meshy2026}, Hunyuan3D~\cite{zhao2025hunyuan3d}, and CLAY~\cite{zhang2024clay} have further broadened access, enabling users to produce textured 3D models from text or image inputs. These generative methods have enabled support tools for creative exploration and rapid prototyping~\cite{shen2024neural}. In addition to generation, a parallel line of work has explored text-driven editing of 3D models (see Lu~\etal~\cite{lu2024advances} for a survey). For example, DreamEditor~\cite{dreameditor2023} enables text-guided editing of 3D scenes by optimizing neural radiance fields, while TIP-Editor~\cite{tipeditor2024} extends this to accept both text and image prompts for localized edits. 

However, while these generative methods produce visually plausible 3D models, they are typically trained using image-based supervision. As a result, they prioritize appearance over physical or functional correctness~\cite{faruqi2024shaping}. This leads to 3D models that may look realistic but lack critical geometric features essential for real-world functionality. Consequently, models generated through these pipelines may exhibit failure modes that only become apparent when used in downstream tasks like simulation or 3D printing. To characterize the scope and prevalence of these flaws, we conducted a formative study examining outputs from a state-of-the-art generative model, which we describe in Section~\ref{sec:formative}.

\subsection{Human-in-the-Loop Customization with Generative Models}

The challenges in generating functional models with generative AI stem from a broader limitation shared across generative modalities. Generative systems must predict complex outputs from underspecified inputs, introducing ambiguity that often leads to outputs misaligned with user intent. Novices struggle particularly with this ambiguity, crafting vague or overly rigid prompts and failing to debug faulty outputs~\cite{zamfirescu2023johnny,subramonyam2023bridging}. This has driven increasing interest in human-in-the-loop approaches that allow users to guide or correct the generative process.

In the image domain, tools such as ControlNet~\cite{zhang2023adding}, PromptPaint~\cite{chung2023promptpaint}, PromptCharm~\cite{wang2024promptcharm}, and Promptify~\cite{brade2023promptify} support iterative refinement through visual conditionings, regional prompting, and attention visualization. Similar trends have emerged with LLMs: ABScribe~\cite{reza2024abscribe}, AI Chains~\cite{wu2022ai}, and EvalLM~\cite{kim2024evallm} enable structured revision, inspectable intermediate steps, and evaluator-guided prompt refinement. \finalAdd{In the 3D domain, part-based composition interfaces similarly give users structural control over generative outputs~\cite{faruqi2026compos3d, faruqi2026mixr}}. Together, these systems illustrate a shift toward mixed-initiative workflows where users and models share creative agency through localized control, visual feedback, and iteration.

InstructMesh extends these principles to 3D generative modeling. Where image-domain tools give users region-level control over appearance, InstructMesh gives users region-level control over fabrication-relevant geometry, enabling novices to correct structural flaws directly on generated models through selection, preview, and iterative revision.

\section{Formative Study} \label{sec:formative}
A key challenge in using generative AI for 3D modeling is ensuring that outputs contain the geometric details needed for physical functionality such as appropriate wall thickness, openings, and manifold integrity. We hypothesize that current generative models replicate aesthetic properties but fail to capture these local geometric details, rendering models non-functional after fabrication. To test this, we conducted a formative study assessing geometric errors in reconstructions of popular 3D models from Thingiverse\footnote{\url{https://www.thingiverse.com}}.


\subsection{Dataset Selection and Generative 3D Reconstruction}
Thingiverse is a popular online resource for novice and expert makers to share designs or things for 3D printing. We selected the 100 most popular `things' (collections of models) from Thingiverse, skipping multi-part components without assembled versions since current generative tools produce single meshes. After preprocessing, the dataset comprised of 120 distinct unique 3D models (100 unique `things' from Thingiverse). We rendered one image of each model and reconstructed it using Trellis~\cite{xiang2024structured3dlatentsscalable}, chosen for its state-of-the-art geometric fidelity. While we scope our analysis on Trellis for consistent reconstruction and evaluation, nearly all current 3D generative methods employ a similar two-stage encoder-decoder pipeline with image-based training objectives.

 \begin{table*}[ht]
\centering
\caption{Taxonomy of Fabrication-Relevant Geometric Flaws in 3D Model Reconstructions. Based on 120 most popular 3D models on 3D model website Thingiverse.}
\label{tab:flaw_taxonomy}
\begin{tabular}{p{4cm} p{9.0cm} >{\raggedleft\arraybackslash}p{2.5cm}}
\toprule
\textbf{Flaw Category} & \textbf{Description} & \textbf{Models Affected} \\
\midrule
Missing Openings & Absence of essential through-holes, ventilation gaps, volumes to contain liquids or support airflow. & 48.6\% \\

Fused Joints & Components meant to rotate or articulate (e.g., hinges, pivots, ball joints, axles) are fused, preventing intended motion. & 13.5\% \\

Broken or Missing Bridges & Crucial geometric connections between parts are missing, leaving the structure fragmented or unsupported. & 10.8\% \\

Extraneous Artifacts & Unintended bumps, extensions, or disconnected mesh elements. & 65.8\% \\

Fused / Duplicated Features & Features are erroneously duplicated, stacked, or fused together, often due to generative ambiguity. & 27\% \\

Hollowing Errors & Solid elements are mistakenly hollowed or vice versa. & 36.9\% \\

Wall Thickness Issues & Non-uniform or insufficient wall thicknesses. & 20.7\% \\

Topological Errors & Gaps, tears, or ambiguous topologies in the mesh. & 9.9\% \\

Truncated Features & Extrusions or geometric extensions are prematurely cut off, resulting in incomplete or malformed shapes. & 9.9\% \\
\bottomrule
\end{tabular}
\end{table*}

\begin{figure}
    \centering
    \includegraphics[width=\linewidth]{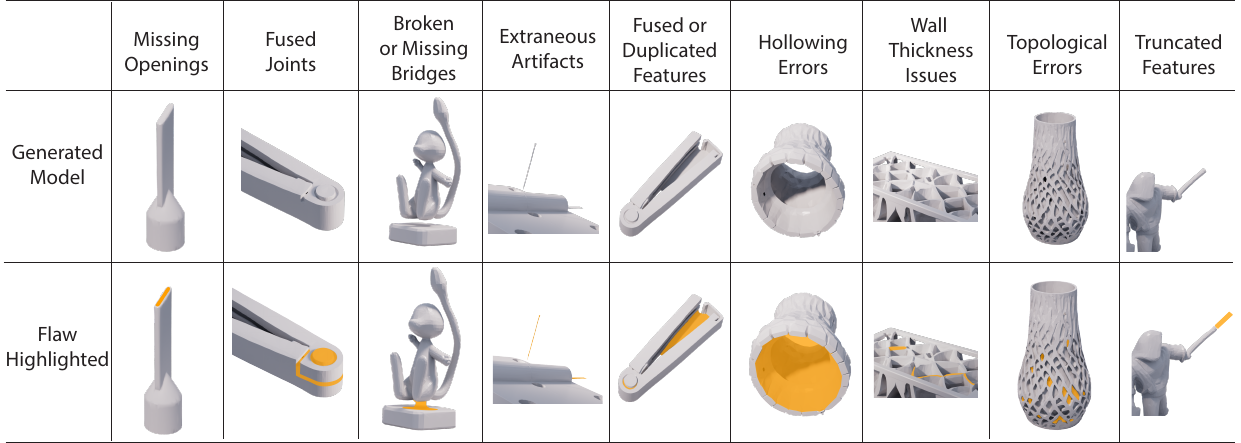}
    \caption{Descriptive examples of flaws found in our formative study. Models were sourced from Thingiverse, reconstructed with 3D generative model Trellis~\cite{xiang2024structured3dlatentsscalable} using their image as input, and evaluated manually for geometric functionality errors by reviewers.}
    \label{fig:flaws_formative}
\end{figure}

\subsection{Inductive Taxonomy Development}
We used an iterative qualitative coding method to develop our taxonomy of geometric flaws in 3D models - geometric errors in reconstruction that impact the functionality of the 3D model. For each of the 120 3D models, two expert annotators in 3D modeling compared the generated model and the original model, and described the issues in natural language. Their task was to describe each individual issue independently, and describe the change required. Next, they discussed these annotations and came to an agreement on the issues found. 

\subsection{Iterative Categorization of Issues}

Next, the two annotators independently grouped similar descriptions of issues, then collaboratively discussed and refined these groupings. We continued this process over multiple rounds of iterative refinement, merging overlapping categories and sharpening definitions through discussion until consensus was reached. The process concluded when no new categories emerged from additional data, at which point the taxonomy had stabilized at nine flaw categories. We report these categories in Table~\ref{tab:flaw_taxonomy} and show descriptive examples in Figure~\ref{fig:flaws_formative}.

\subsection{Deductive Classification}
After developing our taxonomy of fabrication-relevant geometric flaws, we performed a deductive classification of the dataset using the previously collected annotations. The two expert annotators revisited their initial descriptions and mapped each issue to one or more of the nine categories found in the taxonomy. 


Across the 120 models analyzed, we found that fabrication-relevant flaws were widespread: 94 models (78.3\%) exhibited more than one distinct type of issue, highlighting the compounding nature of geometric failures in generative reconstructions. On average, each model exhibited 2.4 issues (SD = 1.10), suggesting that these flaws commonly co-occur and cannot be addressed through isolated fixes. The most frequently encountered issues were Extraneous Artifacts (65.8\%), Missing Openings (48.6\%), and Hollowing Errors (36.9\%), while more localized structural flaws like Broken~/~Missing Bridges (10.8\%) or Truncated Features (9.9\%) appeared less frequently.



These findings suggest that while state-of-the-art generative models achieve high visual fidelity, they overlook fine-grained geometric properties critical for fabrication, which can be difficult to capture through image-based supervision. The prevalence and diversity of these flaws motivate the design of post-generation tools that support targeted, user-guided geometric correction. We describe our approach in the following section.
\section{System Overview}
\label{sec:system}
InstructMesh enables users to generate 3D models with text or image prompts and iteratively refine them to fix fabrication-relevant issues. Recent 3D generative methods, including TRELLIS~\cite{xiang2024structured3dlatentsscalable} and SPAR3D~\cite{huang2025spar3dstablepointawarereconstruction}, employ a two-stage pipeline: an \textit{encoder} produces a coarse latent structure (e.g., point cloud or voxel grid) and contextual features about appearance, and a \textit{decoder} refines this information into a high-fidelity mesh. One key idea is that this separation enables controlled customization by modifying this intermediate representation.  

\begin{figure}
    \centering
    \includegraphics[width=\linewidth]{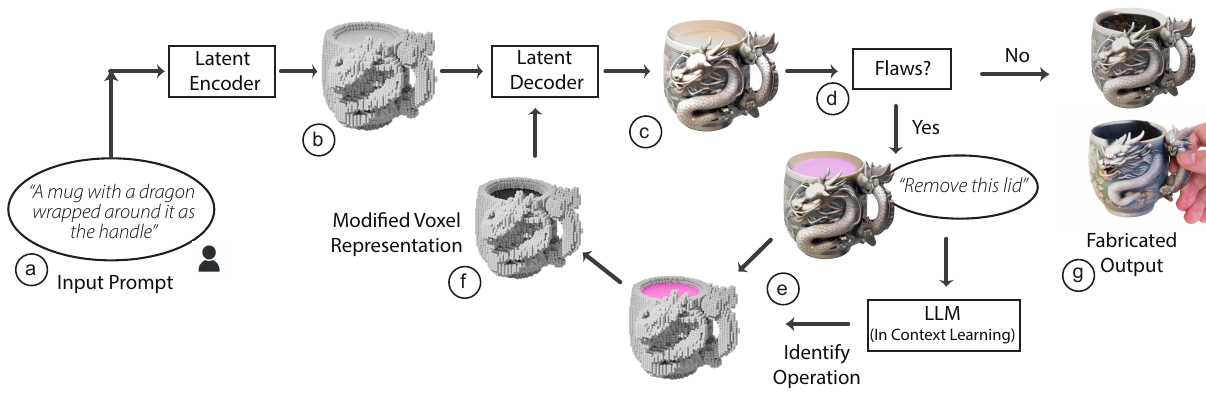}
\caption{InstructMesh System Overview. (a) The user provides a text or image prompt. (b) The model is encoded into a latent voxel representation and (c) decoded into a textured mesh. (d) The user inspects the mesh for flaws; if none, the model proceeds to (g) fabrication. Otherwise, the user highlights a region and specifies an edit. (e) The latent operation is applied and previewed. (f) Once confirmed, the edited latents are decoded into a refined model. }

    \label{fig:system_diagram}
\end{figure}

InstructMesh operates directly on this intermediate latent voxel representation. \finalAdd{We use TRELLIS~\cite{xiang2024structured3dlatentsscalable} as the generative backbone, which represents each shape as a structured latent over a $64^3$ sparse voxel grid; we persist the latent structure alongside the decoded mesh so that user edits can be applied to this latent structure and decoded to produce a new mesh}. Figure~\ref{fig:system_diagram} illustrates the system design, which comprises of: (1) a set of latent-space operations derived from common fabrication flaws, (2) dual interaction modes—a natural language interface powered by an LLM and a slider interface exposing parameterized presets, and (3) a preview visualization that highlights additive edits in green and subtractive edits in red before they are committed. Together, these components provide fine-grained customization with generative 3D models while maintaining user's transparency and control.  

We focus on repairing static functionality in 3D models, as current generative methods do not yet support dynamic mechanisms (e.g., hinges or interlocking parts) that require precise geometric and mechanical constraints. Our goal is to address the static fabrication-related flaws identified in our formative study, with the exception of \emph{Fused Joints}, which involves articulated or mechanical components.  

\begin{figure}

    \centering
    \includegraphics[width=\linewidth]{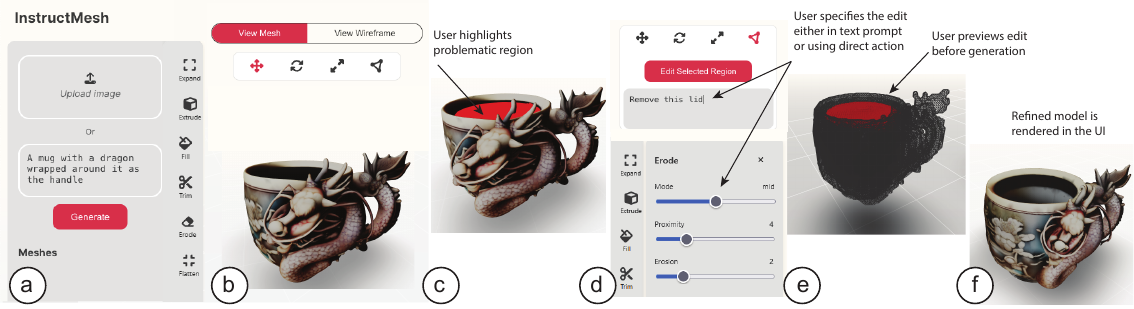}
    \caption{The InstructMesh web interface. (a) Users enter a text or image prompt. (b) The generated mesh is inspected for flaws. (c) Users paint the problematic region. (d) The edit is specified via natural language or direct selection. (e) A preview is shown for confirmation. (f) The refined model is generated.}
    \label{fig:user_interface}
\end{figure}

\begin{figure}
    \centering
    \includegraphics[width=\linewidth]{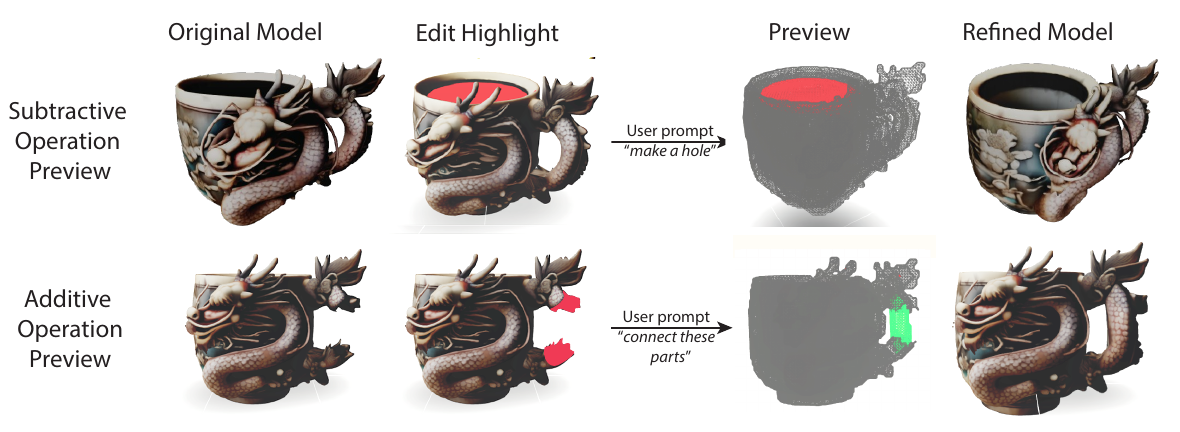}
    \caption{Preview visualizations of canonical operations. Top: a subtractive edit (``make a hole'') removes the sealed lid. Bottom: an additive edit (``connect these parts'') generates a handle. Red regions indicate subtractive changes; green indicates additive. Previews let users verify edits on the latent representation before triggering regeneration.}
    \label{fig:preview_visualization}
\end{figure}

\subsection{User Interface and Workflow}

Figure~\ref{fig:user_interface} shows the InstructMesh web interface. Users begin by providing a text or image prompt to generate a 3D model (Fig.~\ref{fig:user_interface}a), which is rendered for inspection (Fig.~\ref{fig:user_interface}b). If a flaw is identified, they activate \emph{Highlight Region} and paint the problematic area (Fig.~\ref{fig:user_interface}c). Users can then specify the desired edit either by entering a natural language instruction or by selecting an operation from the slider interface (Fig.~\ref{fig:user_interface}d). 

Before the edit is applied, the system presents a \emph{preview} overlay: additive changes are shown in green and subtractive changes in red (Fig.~\ref{fig:preview_visualization}). This makes latent edits transparent and gives users the opportunity to revise or cancel. Once confirmed, the operation is executed on the corresponding subset of the latent representation and decoded into an updated mesh (Fig.~\ref{fig:user_interface}f). The edited model appears next to the original for side-by-side comparison, and users can continue refining, undo an operation, or download the final \finalReplace{fabrication-ready}{repaired} model. 

This workflow enables localized, in-situ corrections, abstracting away the complexity of latent manipulation, while providing transparent feedback loops through preview visualizations. Our goal with this design is to enable novices to iteratively refine generative 3D models with both expressiveness and control.

\subsection{Canonical Operations to Modify Latent Representation}
In this section, we describe our set of latent-space operations that enable localized, precise edits of a generative 3D model. Editing the latent voxel representation, rather than directly modifying the mesh, offers key advantages. Latent grids are coarser and lack explicit connectivity, making them easier to manipulate without navigating complex mesh topology. Additionally, the edited voxels are processed by the decoder, which combines them with learned appearance features to produce a coherent mesh, synthesizing missing geometry without requiring the topology-aware, texture-preserving edits required in tools like Blender. This enables novices to make structural corrections that would otherwise require expert-level mesh manipulation.

\finalAdd{\subsubsection{Mapping Selections to the Latent Representation}
When a user highlights a region on the decoded mesh, the system maps this selection back to the latent voxel grid as follows. First, it recovers the decoder's coordinate transformation by aligning the mesh's bounding box to the voxel grid via scaling, rotation, and translation. The selected faces are then extracted as sub-meshes in this voxel-aligned frame, where they serve as geometric references for the canonical operations: each sub-mesh identifies the voxels to modify. For example, fill uses convex-hull containment, while trim uses signed-distance queries. The modified voxel structure is then passed through the TRELLIS~\cite{xiang2024structured3dlatentsscalable} decoder, which regenerates a coherent mesh from the edited representation.}

\finalAdd{This pipeline is tolerant to imprecise selection through three mechanisms. First, the selection is not treated as an exact voxel mask: connected components below a minimum size are discarded, removing stray fragments from accidental brushing, while all remaining sub-meshes are retained to support disjoint selections (e.g., the two endpoints of a broken bridge). Second, operations act on the neighborhood of these sub-meshes using a proximity radius. Expand, for instance, grows voxels outward from the selection, so that missing a few faces still produces a coherent edit. Third, the preview renders the modified voxels directly, without invoking the decoder, allowing users to verify the actual effect of an operation before committing (Fig.~\ref{fig:preview_visualization}). We provide pseudocode in Appendix A.}

\begin{figure}
    \centering
    \includegraphics[width=\linewidth]{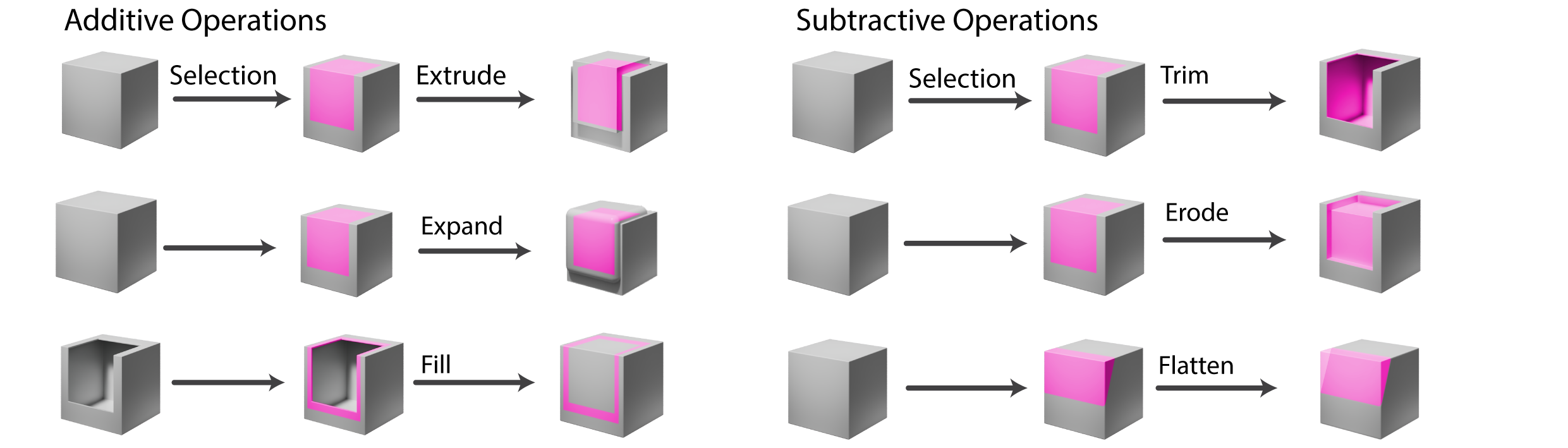}
    \caption{Canonical operations on voxel representations. User highlights (pink) are mapped to regions in the latent voxel grid. Additive operations (left) grow occupancy via extrusion, expansion, or filling. Subtractive operations (right) reduce occupancy via trimming, erosion, or flattening. The modified grid is decoded to produce the refined model.}
    \label{fig:operations}
\end{figure}

\finalAdd{With selections mapped onto the latent grid, the system next needs a vocabulary of edits to apply there.} We defined a small set of \textit{canonical} operations analogous to geometric primitives such as extrude or subtract, that are compatible with latent voxel grids and composable: individual edits can be combined or sequenced to address complex repair scenarios (Fig.~\ref{fig:operations}). \finalDel{When users highlight a flawed region on the mesh, the system maps this region to the latent representation and applies one of these operations before decoding back into a refined mesh.}

We group our operations into two classes:
\subsubsection{Additive Operations}
These increase voxel occupancy by extending surfaces outward, thickening thin regions, or filling enclosed volumes. They are used to reinforce fragile areas or close gaps.
\begin{itemize}
    \item \textbf{Extrude} extends the selected surface region by adding voxels along its average outward normal, useful for lengthening handles or creating protrusions. 
    \item \textbf{Expand} thickens structure by adding a uniform shell of voxels around the selection, reinforcing thin walls and fragile geometry.
    \item \textbf{Fill} closes surface gaps or hollow volumes by populating voxels inside the convex hull of the selection, commonly used to solidify hollow parts or create bases and lids.
\end{itemize} 

\subsubsection{Subtractive Operations}
These reduce occupancy by cutting away selected regions, thinning shells, or smoothing surfaces. They are used to remove unwanted material, hollow interiors, or prepare flat contact areas.
\begin{itemize}
    \item \textbf{Trim} removes voxels near the selected surface based on a signed distance field threshold, useful for cleaning artifacts or cutting away extraneous material.
    \item \textbf{Erode} thins a surface or creates voids by subtracting an interior shell of voxels, often used to hollow parts or carve channels. 
    \item \textbf{Flatten} smooths a region toward a best-fit plane, preparing contact surfaces or flat bases for fabrication.
\end{itemize}


Together, these operations form a compact, extensible vocabulary for repairing generated 3D models.\finalDel{Internally, these operations dynamically select among voxel grids, surface meshes, and signed distance fields depending on the task. For example, \textit{trim} uses SDF-based distance queries while \textit{fill} computes convex hulls over the mesh.} The latent voxel representation provides a meaningful locus of operations, and the flexibility of decoding process into the final mesh allows us to create these representation-agnostic operations. Thus, these representation switches are abstracted away from the user who can focus on the high level design intent. Each operation can be parameterized (e.g., extrusion depth, erosion thickness), enabling a range of functional modifications. Formal definitions and pseudocode for all operations are provided in Appendix~A.

\subsubsection{Operation Prediction Using LLMs with In-Context Learning}
\label{sec:LLM_icl}

To map natural language edit descriptions to canonical operations, we employ in-context learning (ICL) with OpenAI's GPT-4~\cite{gpt4techreport}. \finalAdd{The LLM does not interpret or edit 3D geometry directly: it maps the user's instruction to an operation and parameters from our fixed vocabulary, while spatial grounding comes from the user's region selection. Constraining the LLM to select from a pre-parameterized set of operations rather than synthesize direct edits, keeps each edit interpretable and bounded. This is consistent with prior work showing that scoping an LLM to a fixed set of operations improves reliability in human-AI interaction~\cite{wu2022ai}}. We provide a set of labeled description-operation pairs drawn from the 20\% development split of our formative dataset (see Section~\ref{sec:tech-eval}) directly in the prompt. For example, given the description ``the mug has a sealed lid that should be open,'' the model predicts \textit{erode} as the appropriate operation. By including examples spanning all operation types and flaw categories, the model learns to map new user instructions to the fixed operation vocabulary without fine-tuning. We chose ICL over fine-tuning for its simplicity and adaptability: new operations can be incorporated by extending the example set without retraining. While our implementation uses GPT-4, the approach is model-agnostic: any modern LLM model can serve as the backend with the proposed ICL approach. We evaluate prediction accuracy in Section~\ref{sec:tech-eval}.

\section{Evaluating and Calibrating Operations}
\label{sec:tech-eval}

In the previous section, we introduced a set of canonical operations for making fine-grained edits to the latent voxel representation prior to mesh regeneration. Each operation has tunable parameters that must be calibrated to reliably rectify the flaws identified in our taxonomy. To this end, we conducted a technical evaluation. We first randomize our dataset, and split it using a 80\%-20\% split. We used 20\% set as a `development set' to calibrate each operation, and use the 80\% to evaluate its accuracy in resolving an issue. 

Using the development set, we manually calibrated each operation (e.g., \emph{expand}, \emph{extrude}, \emph{erode}, \emph{fill}) by highlighting affected regions and testing parameter ranges for their ability to resolve the target flaw. We found that different flaw contexts required distinct parameterizations. \finalReplace{To cater to diverse editing scenarios while controlling complexity, we defined a limited set of calibrated presets. Too many presets would increase interface complexity for novices and affect LLM prediction accuracy, whereas too few would reduce flexibility. We therefore defined}{Balancing flexibility against interface complexity and LLM prediction accuracy, we defined} twelve calibrated presets across six operations. These presets function as operational modes that can be invoked via natural language or the slider interface; detailed descriptions are provided in Appendix~B.

\subsection{Technical Evaluation}

\subsubsection{Flaw Correction Accuracy:}
We tested our calibrated operations on the 80\% test split from our formative study. The operations were conducted by the authors with InstructMesh, and the resulting 3D models were stored. To evaluate the repairs, an independent reviewer assessed each flaw and judged whether the correction was successful or unsuccessful. This evaluation was conducted at the flaw level, meaning each flaw was independently repaired, stored, and assessed by the independent reviewer. 

\begin{figure}
    \centering
    \includegraphics[width=0.8\linewidth]{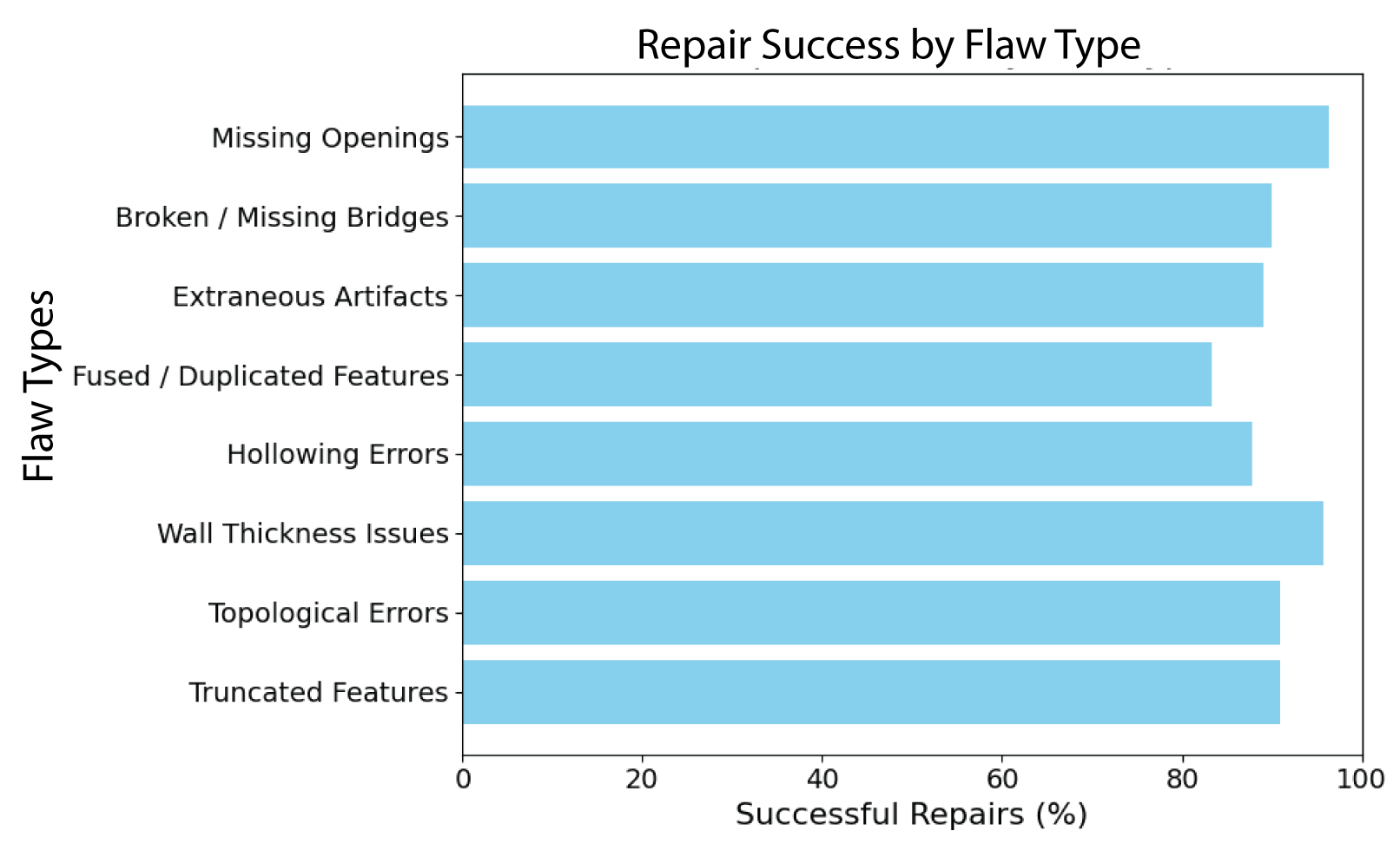}
    \caption{Success rates of canonical operations in resolving fabrication-relevant geometric flaws. This evaluation was conducted on 80\% of flaw instances from our annotated dataset. For each flaw, a predefined canonical operation was applied through region selection and LLM-guided latent editing.}
    \label{fig:technical_eval}
\end{figure}

As shown in Figure~\ref{fig:technical_eval}, InstructMesh successfully repaired a high proportion of flaws across all categories. The most reliably repaired issues included Missing Openings (96.3\%) and Wall Thickness Issues (95.65\%), while the lowest success rate was observed for Fused / Duplicated Features at 83.33\%. These results demonstrate that the latent operations implemented in InstructMesh are broadly effective across a diverse set of fabrication-relevant flaw types.

\subsubsection{Runtime Efficiency:} 
We also evaluated runtime performance. Using Trellis for image-to-3D generation, the average generation time was 31.7s (SD = 3.7s). Applying region-specific latent operations increased this to 43.7s (SD = 5.2s), reflecting the added computation for localized manipulation and regeneration. Despite this overhead, edits remained well under a minute, supporting iterative design workflows.


\subsubsection{LLM Accuracy for Operation Selection:}
Using the ICL setup described in Section~\ref{sec:LLM_icl}, GPT-4 achieved 92.1\% accuracy in predicting the correct canonical operation on the 80\% test split, demonstrating that in-context learning is sufficient to map natural language instructions to our operation vocabulary without fine-tuning.

\subsubsection{Implementation Details:} 
The generative pipeline is implemented in Python and deployed on a cloud-based server (Intel CPU, 32 GB RAM, NVIDIA L4 GPU with 24 GB memory). Canonical latent-space operations were implemented using PyTorch, \texttt{Pymeshlab}~\cite{pymeshlab} and \texttt{Trimesh}~\cite{trimesh2019}.

\section{User Evaluation I: Novice Flaw Identification and Repair}

For InstructMesh to be effective, novices must be able to (1) identify functional flaws in generated models and (2) repair them independently and efficiently using our system. We designed our first study to evaluate both capabilities.

\subsection{Study Design}
\begin{figure}
    \centering
    \includegraphics[width=\linewidth]{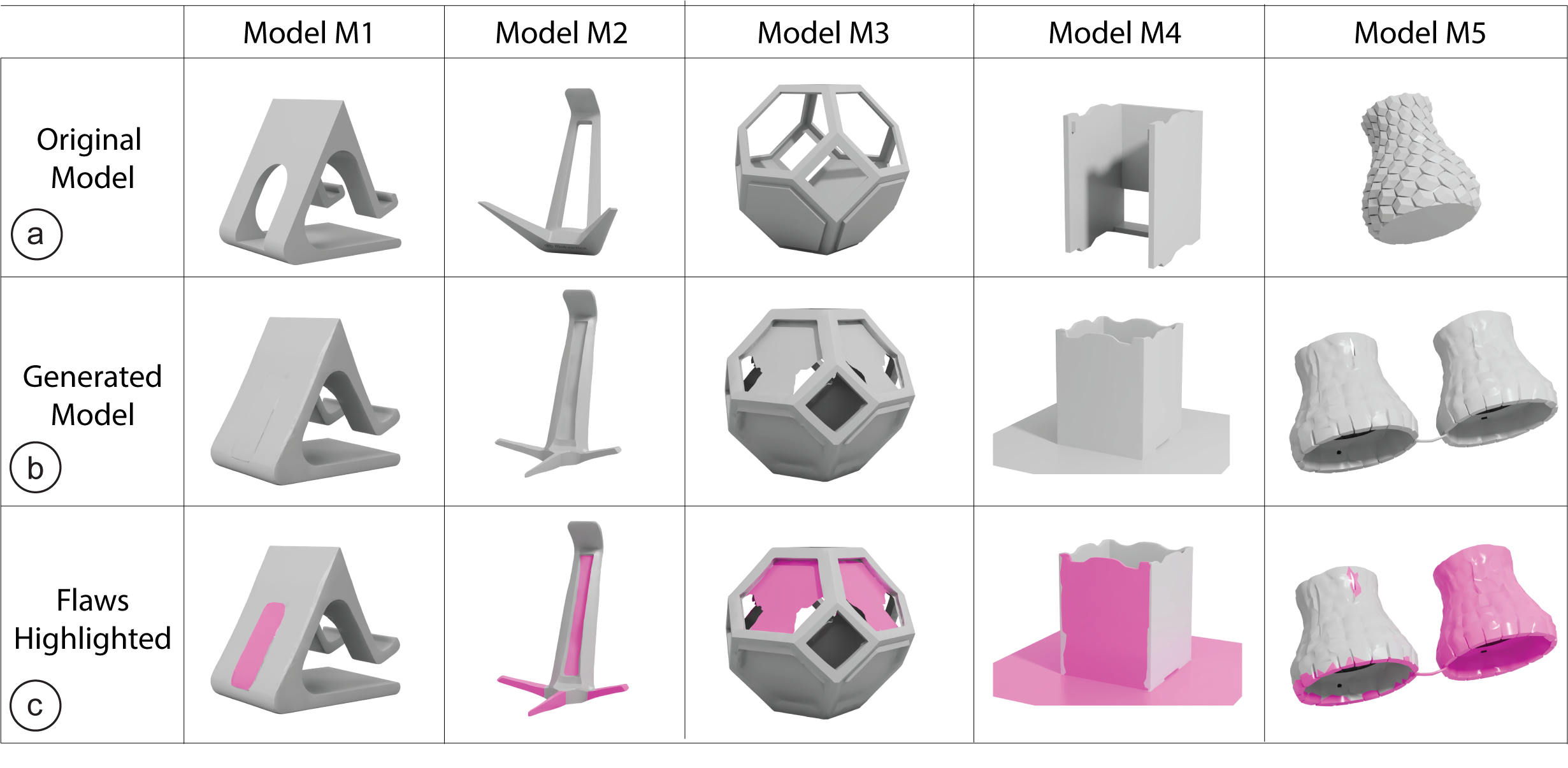}
    \caption{Overview of the five models used in our user study (M1–M5). (a) Original model. (b) Flawed version from a generative model. (c) Fabrication-related flaws highlighted in pink. }
    \label{fig:user_study_models}
\end{figure}

We recruited 12 participants (7 male, 5 female; ages 20–39, $M=27.3$, $SD=5.6$) from a university campus, including undergraduate and graduate students across non-technical disciplines. None had prior experience with 3D modeling, CAD tools, or digital fabrication. Participants worked with five flawed 3D models randomly drawn from our formative study (Fig.~\ref{fig:user_study_models}), each containing between one and four annotated fabrication flaws.
For each model, participants first inspected the mesh and identified flaws while thinking aloud. They then used InstructMesh to attempt repairs, with up to 10 minutes per model. Participants could address flaws in any order. After completing their edits, participants judged for each flaw whether the repair was successful or unsuccessful. An independent researcher, post user-study sessions, evaluated the final meshes in the same way. We measured task completion time, the number of flaws identified and repaired, and repair success rates based on both participant and expert judgments.


\subsection{Results}

\begin{figure}
    \centering
    \includegraphics[width=\linewidth]{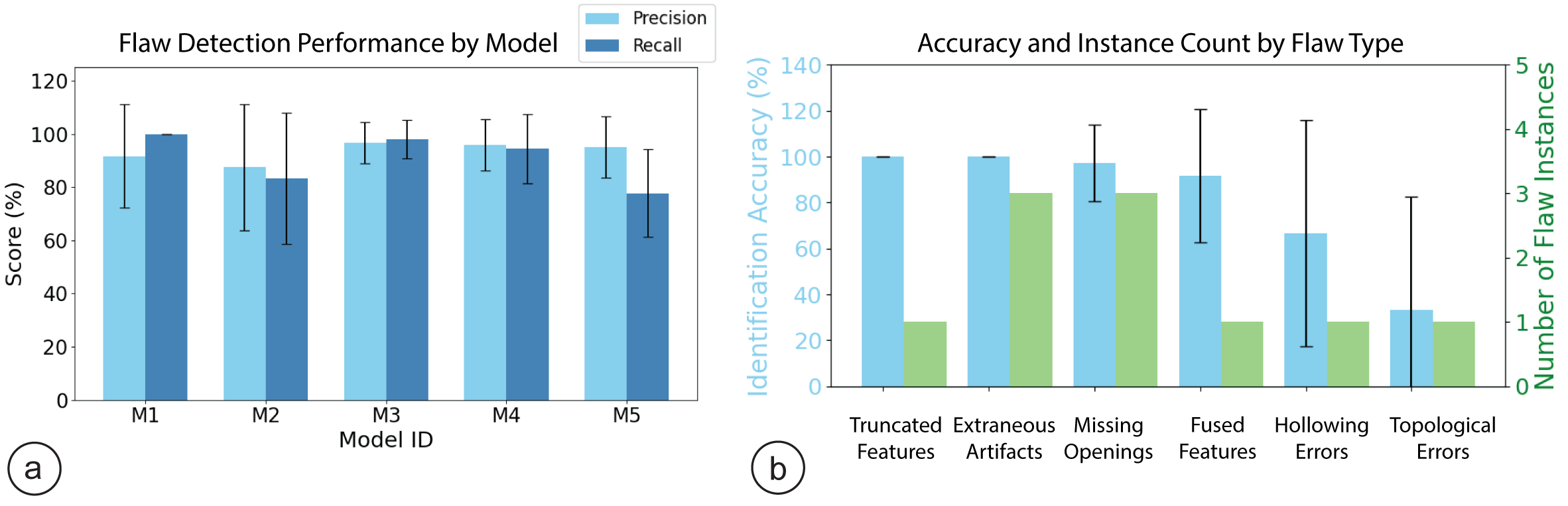}
    \vspace{-5mm}
    \caption{Flaw Identification Performance.
(a) Precision and recall of flaw detection across five generated 3D models (M1–M5). (b)~Average identification accuracy (left axis) and number of flaw instances (right axis) for each flaw type. }
    \label{fig:user_study_identification-1}
\end{figure}

\begin{figure}
    \centering
    \includegraphics[width=\linewidth]{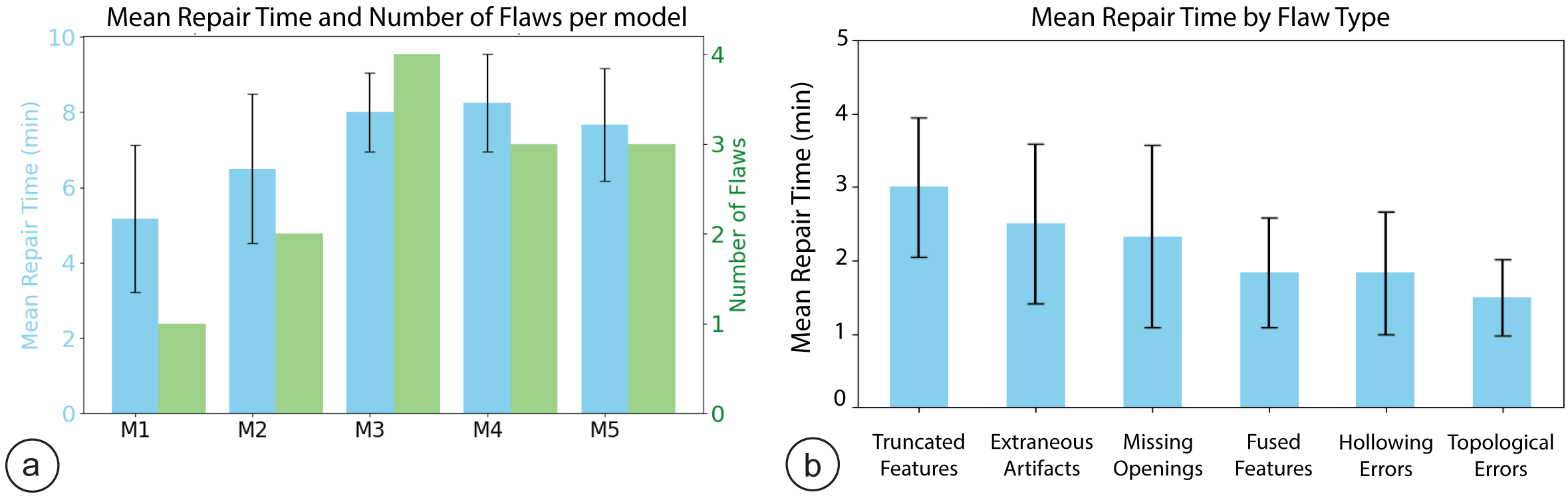}
    \caption{Repair Time and Effort.
(a) Mean repair time and number of flaws per model, indicating that models with more flaws required longer repair times.
(b) Mean repair time by flaw type. Simpler flaws like topological errors and fused features were repaired faster, while truncated features and missing openings took longer.}
    \label{fig:user_study_repair}
\end{figure}

\begin{figure}
    \centering
    \includegraphics[width=0.6\linewidth]{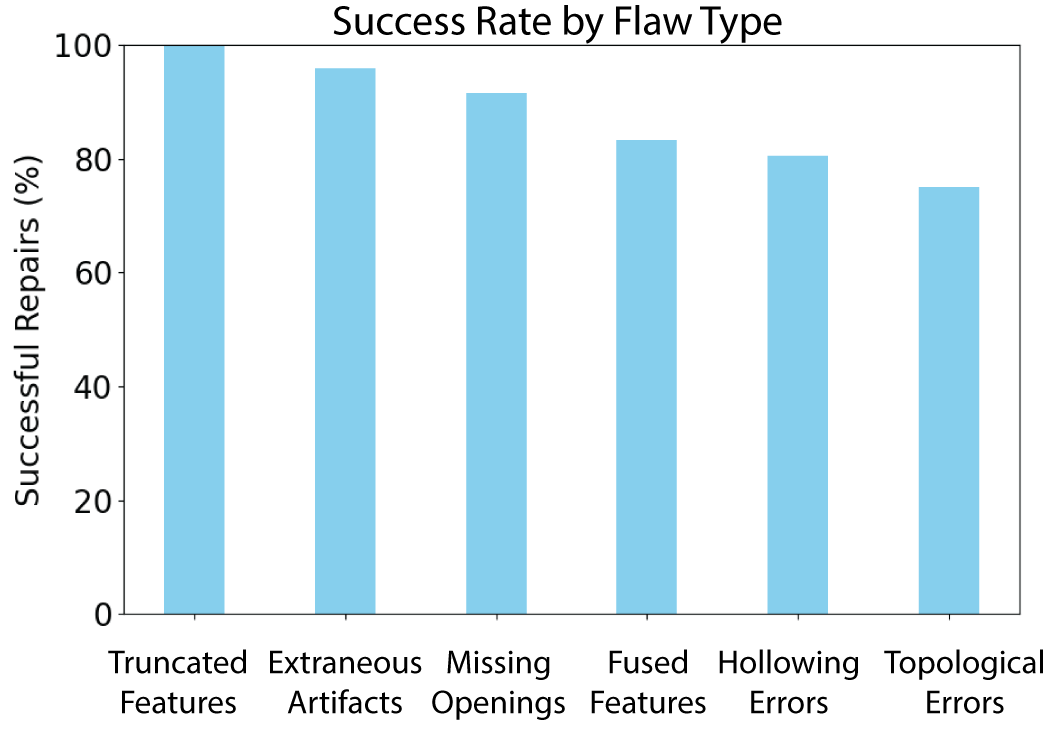}
    \caption{Repair Success by Flaw Type.
Participants successfully repaired most flaw types, with success rates exceeding 80\% for all categories.}
    \vspace{-5mm}
    \label{fig:user_study_sucess}
\end{figure}

We report findings across flaw identification, repair time, and repair success, using repeated-measures ANOVAs to test for effects of model and flaw type.

\subsubsection{Error Identification}

Participants successfully identified 90.4\% of fabrication-relevant flaws in AI-generated models (Fig.~\ref{fig:user_study_identification-1}), showing that novices could reliably recognize most issues. However, performance varied by both model and flaw type. A repeated-measures ANOVA revealed a significant effect of model (\(F(4,55)=3.06, p~<~0.05\)); post-hoc tests showed that flaws in Model M1 were significantly easier to detect than those in Model M5 (\(p<0.05\)). This suggests that flaws in more geometrically complex models can be harder to perceive.  

Flaw type also had a significant effect (\(F(5,139)=19.24, p~<~0.0001\)). Topological issues were the hardest to detect, performing significantly worse than truncated features, extraneous artifacts, missing openings, and fused features. In contrast, extraneous artifacts and missing openings were among the most readily identified. Together, these results demonstrate that while novices can generally identify fabrication flaws, both the geometry of the model and the type of flaw strongly affect detectability.

\subsubsection{Repair Times}

Repair efficiency varied significantly across both models and flaw types (Fig.~\ref{fig:user_study_repair}). A repeated-measures ANOVA revealed a significant effect of model on repair time (\(F(4,55)=7.27, p<0.001\)), with simpler models (e.g., M1) repaired significantly faster than more complex ones (M3, M4, M5).


Flaw type also had a significant effect on repair time (\(F(5,121)=5.43, p<0.001\)). Truncated features (3.00 ± 0.95 min) took significantly longer to repair than topological errors (1.50 ± 0.52 min, \(p<0.01\)) and fused features (1.84 ± 0.75 min, \(p<0.05\)); missing openings (2.33 ± 1.24 min) also took longer than topological errors (\(p<0.05\)). These results suggest that while flaws such as topological errors were harder to identify, they could be resolved quickly once recognized, whereas flaws like truncated features required more time-consuming interventions.

\subsubsection{Repair Correction Performance}

An independent expert in 3D modeling and fabrication evaluated all user-corrected meshes (Fig.~\ref{fig:user_study_sucess}). Each flaw was judged as successfully repaired or not, based on whether functionality was restored. Overall, 89.7\% of flaws were successfully repaired, with only 16 failures out of 156 annotated flaws.  

A repeated-measures ANOVA showed no significant effect of model on repair correctness (\(F(4,55)=2.25, p>0.05\)), but flaw type had a significant effect (\(F(5,121)=2.33, p<0.05\)). Post-hoc tests, however, did not reveal significant differences between specific flaw categories. This suggests that while some flaw types may be more challenging, InstructMesh enabled novices to produce \finalReplace{fabrication-ready}{successful fabrication-relevant} repairs across diverse models.  

We also compared participants’ self-assessments with the expert’s ratings. No significant difference was found (\(F(1,587)=0.40, p>0.05\)), indicating close alignment between novices’ perception of successful repair and expert evaluation. Only 11 cases (7.1\%) showed disagreement, where participants believed a flaw was fixed but the reviewer did not.

Overall, these results suggest that novice users were able to identify and correct functional errors in AI-generated 3D models with a high degree of accuracy. 

\subsection{Participant Feedback}
Participants responded positively to InstructMesh, highlighting its ease of use and low barrier to entry. P2 remarked, ``\textit{I like that you can undo your changes, and I just have to describe my changes in text. It's very easy to get used to the tool.}'' P6 emphasized that ``\textit{I don't feel the need to learn CAD software,}'' while another noted ``\textit{First time 3D modeling - Didn't know it could be easy!}'' Participants envisioned use cases spanning rapid prototyping, creative hobbyist printing, and educational contexts.

\section{User Evaluation II: Interaction Modes and Preview Visualization}
\label{sec:comparative_study}

Our first user study demonstrated that novices can identify and repair fabrication-relevant flaws using InstructMesh. The second study examines how the choice of \textit{interaction mode} shapes this experience, comparing two conditions: (1) an LLM-based interface where participants describe edits in natural language, and (2) a slider-based interface where participants select operations directly. We also investigate the role of \textit{preview visualization} in supporting user understanding and clarity during latent-space editing.


We address the following research questions:
\begin{itemize}
    \item \textbf{RQ1 (Usability \& Control):} How do the LLM and slider interfaces differ in terms of ease of use, effort, clarity of intent expression, perceived accuracy, and sense of control?
    \item \textbf{RQ2 (Previews):} How do latent-space previews influence participants’ understanding of edits, their confidence, and their decision to accept or revise?
    \item \textbf{RQ3 (Trade-offs):} Do participants prefer natural language versus direct operation selection, or a combination? 
\end{itemize}

\subsection{Tasks and Procedure}

We recruited 12 new participants (7 female, 5 male; compensated 20~USD), none with prior 3D modeling or CAD experience. Each completed 12 editing tasks from the same flawed models used in Study~1, split into two counterbalanced blocks of six: one with the LLM interface and one with sliders.

For each task, participants were shown a flawed model and its target corrected version, and asked to rectify the error using the assigned interface. In the LLM condition, participants described desired edits in natural language, while in the slider condition, they selected operations and adjusted parameters directly. In both conditions, a preview visualization was shown before committing the edit.

After each block, participants completed six 7-point Likert items assessing ease of use, perceived accuracy, clarity of intent, sense of control, effort, and preview usefulness. This evaluation was informed by usability constructs from standardized questionnaires such as SUS~\cite{brooke1996sus}, adapted to our study-specific tasks. A post-study questionnaire asked participants to compare interfaces and indicate preference for LLMs, sliders, or a hybrid workflow.

\subsection{Results}

\begin{figure}
    \centering
    \includegraphics[width=\linewidth]{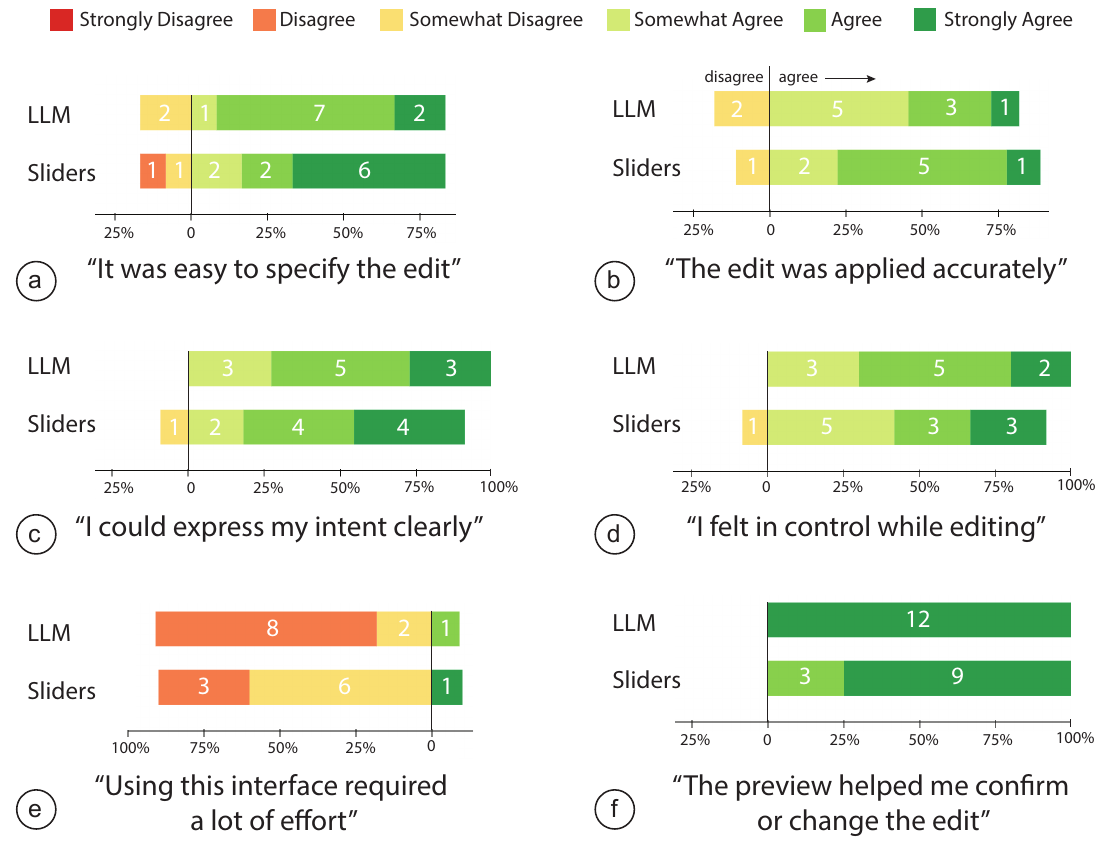}
    \caption{Likert results comparing LLM and slider interfaces. (a,c) LLM rated easier for specifying edits. (b,d) Sliders scored higher on accuracy and control. (e) LLMs required less effort. (f) Previews rated very highly across both conditions. }
    \label{fig:user_study_identification}
\end{figure}

We analyzed the questionnaire data using descriptive statistics and within-subject comparisons between the LLM and slider conditions. Figure~\ref{fig:user_study_identification} shows the results from the analysis. 

\paragraph{RQ1 (Usability \& Control).}  

Both interfaces were rated as usable, with subtle differences. The LLM interface was rated as less effortful (\textit{M} = 2.67, \textit{SD} = 1.23 vs. \textit{M} = 3.25, \textit{SD} = 1.36), while the slider interface was perceived as slightly more accurate (\textit{M} = 5.17, \textit{SD} = 1.19 vs. \textit{M} = 5.00, \textit{SD} = 1.21). Perceived control was comparable (both \textit{M} = 5.58). This suggests a trade-off: LLMs lower the barrier to entry, while sliders provide a stronger sense of precision.

\paragraph{RQ2 (Previews).}  

Previews received consistently high ratings across both interfaces. Participants agreed the preview clearly showed what the system would do (\textit{LLM: M = 6.67, SD = 0.65; sliders: M = 6.75, SD = 0.45}) and helped them decide whether to accept or revise (\textit{LLM: all participants rated 7.0; sliders: M = 6.75, SD = 0.45}). One participant noted the preview ``\textit{made invisible operations visible},'' suggesting that previews are critical for building trust in non-WYSIWYG editing.

\paragraph{RQ3 (Trade-offs).}  
Participants were evenly split in overall preference (\textit{6 sliders, 6 LLMs}), and all 12 favored a hybrid design combining both modes. Users preferring the slider mode emphasized precision: \textit{``I had more control of the edits rather than waiting to see if my intent was clearly understood by the LLM''} (P1). Meanwhile, LLM advocates highlighted expressiveness: \textit{``It was easier to specify the modification I wanted''} (P6). Participants described the two modes as complementary, natural language for broad creative edits such as \textit{``generating initial sketches and making big fundamental changes''} (P4), and sliders for precise local refinements such as \textit{``fill in a gap, make a hole''} (P2). These findings suggest that novices value the expressiveness of LLMs and the precision of sliders, and see the hybrid combination with previews as the most effective design direction.

\section{Applications}

We demonstrate InstructMesh across six application scenarios covering home decor, medical devices, personal accessories, and robotic enclosures. All examples were constructed with InstructMesh and printed on a Stratasys J55 multi-color 3D printer.

\begin{figure*}
    \centering
    \includegraphics[width=\linewidth]{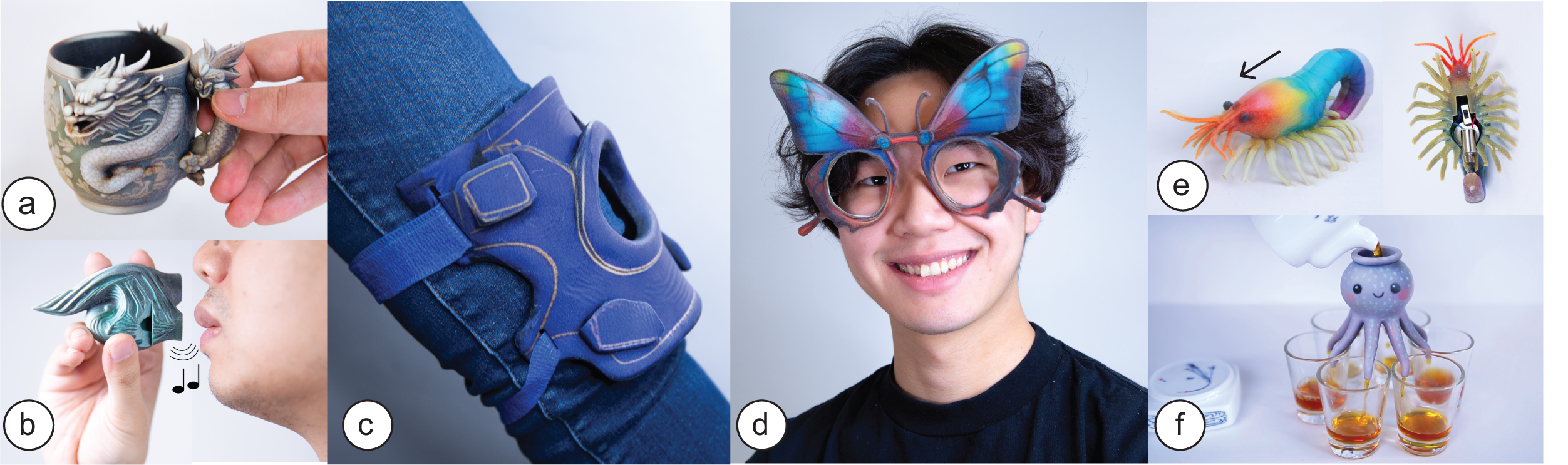}

    \caption{Applications of InstructMesh across domains. (a) Dragon mug with lid removed. (b) A functional whistle with acoustic chambers added. (c) Denim-aesthetic knee brace with strap openings added. (d) Butterfly eyeglasses with hollowed lenses and connected temples. (e) Bristle bot enclosure with electronics cavity. (f) Octopus-inspired liquids dispenser with added fluid channels.}
    \label{fig:applications}
\end{figure*}



\subsection{Home Decor and Functional Objects}
Personal fabrication tools are increasingly used to design expressive yet functional home objects. We showcase a mug with a dragon-shaped handle, where the generated model included a sealed lid. With InstructMesh, the lid was removed, producing a \finalReplace{fabrication-ready}{printable, functional} version (Fig.~\ref{fig:applications}a). We also demonstrate an octopus-inspired liquid dispenser whose head and legs were sealed. We created a top opening, thickened the surrounding area, and added tunnels through each leg to enable fluid flow (Fig.~\ref{fig:applications}f).



\subsection{Medical and Assistive Devices}
Personalized fabrication is increasingly important in ``Medical Making''~\cite{Lakshmi_poc} and DIY Assistive Technology~\cite{buehler2015sharing}, where combining function and aesthetics improves adoption~\cite{Shinohara_social}. We generated \textit{``a knee brace in the style of blue denim jeans''}. The model achieved the desired aesthetic but lacked functional openings for straps. Using localized edits, we created strap holes to produce a wearable, \finalReplace{fabrication-ready}{functional} design, \finalAdd{ready for printing}~(Fig.~\ref{fig:applications}c).

\subsection{Personal Accessories}
InstructMesh also supports the refinement of personalized accessories. We generated \textit{``butterfly-shaped eyeglass frames''} with ornate wing patterns. The initial model had filled lens regions, a thin bridge, and disconnected temples. Localized edits corrected these flaws into a functional design (Fig.~\ref{fig:applications}d). We also generated \textit{``an ornate whistle in the shape of a seashell''} that lacked acoustic chambers. Using InstructMesh, we added tunnels and refined the resonant geometry; the fabricated whistle produces a clear tone (Fig.~\ref{fig:applications}b).

\subsection{Robotic Enclosures}

Finally, we demonstrate InstructMesh’s utility in robotics with a shrimp-shaped enclosure for a bristle bot~\cite{zhu2020curveboards}. We generated a \textit{``colorful shrimp with several legs''} as an enclosure for a bristle bot housing a DC motor. The model's front and back legs would cancel forward motion, and it lacked a cavity for electronics. We shortened obstructing legs, created an underbelly cavity, and added a motor tunnel, producing a \finalReplace{fabrication-ready}{printable} enclosure that enabled directed walking (Fig.~\ref{fig:applications}e).

\section{Discussion and Future Work}

InstructMesh demonstrates how fabrication-aware editing can be integrated into generative 3D workflows through selective highlights and natural language interactions. In this section, we discuss the broader implications of our system design and study results, the current limitations, and future directions. 


\subsection{Insights and Design Guidelines}

\finalDel{Our two user studies reveal several patterns that inform the design of future fabrication-aware generative tools.} In Study~1, novices identified 90.4\% of fabrication-relevant flaws and successfully repaired 89.7\%, suggesting that post-generation repair can be made accessible through latent-space operations paired with region-based selection. This aligns with prior observations that novice makers can reason about physical properties of 3D models but lack the expert skills with 3D~modeling tools to perform the relevant corrective operations~\cite{hudson2016understanding, oehlberg2015patterns}. Across flaw types, detectability and repair difficulty were not always correlated (Figures~\ref{fig:user_study_identification-1},~\ref{fig:user_study_repair}), suggesting that future tools could benefit from both automated flaw detection and operation recommendations.

In Study~2, the even split between LLM and slider preference (6/6), combined with unanimous preference for a hybrid workflow (12/12), provides empirical support for dual-mode interaction design. \finalDel{LLMs reduced perceived effort while sliders increased perceived accuracy, highlighting a trade-off consistent with findings in other mixed-initiative systems. The unanimously high preview ratings underscore that when the underlying representation is non-visual, previews are essential for user trust and decision-making.} From these findings, we distill four design guidelines:
\begin{itemize}
    \item \textbf{Offer complementary input modes.} Natural language lowers the barrier to entry; structured controls provide precision. Supporting both lets users match the mode to the task.
    \item \textbf{Provide previews for non-WYSIWYG edits.} Latent-space operations are inherently opaque. Visual previews make edits tangible, improving transparency and trust.
    \item \textbf{Bridge intent and operation with parameterization.} Calibrated presets make operations reliable while leaving room for adjustment, reducing cognitive load without oversimplifying.
    \item \textbf{Prioritize near real-time feedback.} Edits and previews that complete within seconds are sufficient to maintain iterative workflows and user engagement.
\end{itemize}

\subsection{Heuristics vs. Learned Approaches for Latent-Space Editing}
We opted for heuristic-based operational design for several reasons. First, they are interpretable: each operation has a clear geometric meaning (e.g., ``remove voxels within a distance threshold'') that can be previewed, parameterized, and explained to users. This interpretability is central to our WYSIWYG design goal. Second, they are composable: simple operations can be sequenced to address complex repairs, as demonstrated by combinations like \textit{fill} + \textit{erode} for creating bridges. Third, they require no training data, which is important given the absence of large-scale datasets of fabrication-relevant 3D edits.

A learning-based approach, such as training a model to predict voxel modifications from natural language descriptions, could potentially handle a broader range of edits and generalize to novel repair scenarios. However, such an approach could limit the transparency and controllability that our user studies identified as critical for novices. We see the two approaches as complementary: heuristic operations provide a reliable, inspectable foundation, while learned components (such as our ICL-based operation prediction) can route user intent to the appropriate operation. Future work could explore hybrid architectures where learned models propose edits that are executed through interpretable geometric primitives.

\finalAdd{Although our implementation builds on TRELLIS, the approach is generalizable across similar two-stage generative backbones. The canonical operations act on a structured intermediate voxel representation, including recent models such as TRELLIS~2~\cite{xiang2026native} and SAM3D~\cite{chen2026sam}. Similarly, the ICL-based operation mapping is model-independent: we validated it with GPT-4, but the same in-context approach can route instructions to operations with any modern LLM. Our formative dataset, drawn from the most popular Thingiverse designs, reflects the contexts in which novices most commonly fabricate objects, and broader fabrication domains are important areas of future work.}

\subsection{Limitations}
\label{sec:limitations}
We identify several limitations and future directions for this work. We focused our evaluation on a single generative model, Trellis, due to its state-of-the-art reconstruction fidelity and to maintain a standardized evaluation pipeline. We \finalReplace{plan to}{will} open-source our implementation to support future analysis across additional generative models as they continue to evolve.

\finalAdd{The $64^3$ voxel grid bounds the granularity of repairs: a single voxel spans roughly 1.5\% of a shape's extent, which sets the smallest feature an edit can target. The flaws identified in our formative study are meso-scale and span many voxels, making them addressable, as our evaluation demonstrates. However, sub-voxel defects cannot be repaired in the current implementation. This is a limitation of grid resolution rather than of the operations themselves and higher-fidelity backbones can improve the minimum addressable feature size.}

\finalAdd{InstructMesh supports the repair of fabrication-relevant flaws but relies on the user to diagnose them: the system assumes users can visually identify and select a problematic region. Our studies show novices do this reliably for visually apparent flaws (90.4\%, Study 1), but defects such as internal wall-thickness violations or tolerance issues may only surface during slicing or after a failed print. Automatic flaw detection and operation recommendation are natural and exciting future work.}

Our user studies were designed to evaluate whether novices can identify and rectify \finalAdd{visually identifiable} fabrication-relevant flaws within a generative workflow. \finalDel{This question was motivated directly by the prevalence of geometric errors found in our formative study. Study~1 assessed the feasibility of this task, and Study~2 examined how different interaction modes shape the editing experience.} \finalReplace{A comparative evaluation against}{We did not compare InstructMesh with} alternative workflows such as iterative re-prompting or manual mesh repair. \finalAdd{Such analysis} would complement\finalDel{these} findings \finalAdd{from this paper} and is an important direction for future work. \finalAdd{Similarly, our fabricated examples demonstrate applications rather than systematic printability validation.}

\finalReplace{We limited the scope of InstructMesh to static geometry since current generative models create static geometry. Thus this current system cannot address flaws involving dynamic mechanisms such as hinges, or interlocking parts.}{As scoped in Section~\ref{sec:system}, InstructMesh targets static geometry and cannot address flaws involving dynamic mechanisms such as hinges or interlocking parts.} As these generative systems continue to evolve, future systems can incorporate dynamic simulations to evaluate functionality. Finally, the preview visualization uses red and green overlays, which may pose challenges for colorblind users; future iterations should explore alternative visual encodings such as patterns or shape cues~\cite{zhang2025a11yshape}.

\section{Conclusion}
In this work, we present InstructMesh, \finalReplace{a system}{an interactive post-generation refinement tool} that enables novices to \finalReplace{refine}{perform fabrication-relevant repairs on} generative 3D models \finalDel{for fabrication} by mapping region selections and natural language instructions to canonical latent-space operations. A formative study revealed geometric flaws in state-of-the-art generative outputs, motivating a set of composable editing operations that can be triggered via text or slider controls. A preview visualization makes this latent-editing process transparent, supporting user trust and decision-making. Our user studies show that novices can detect and repair \finalAdd{visually identifiable} flaws in under ten minutes per model, and that hybrid workflows combining natural language with structured controls are preferred. We demonstrate practical applications spanning home decor, assistive devices, personal accessories, and robotics.

\begin{acks}
We would like to extend our sincere gratitude to the MIT-Google Program for Computing Innovation, the MIT-HPI Collaborative Research Program, and the MIT Felicis Program for their generous support, which made this research possible.
\end{acks}

\bibliographystyle{ACM-Reference-Format}
\bibliography{references}

@article{hong2023lrm,
  title={Lrm: Large reconstruction model for single image to 3d},
  author={Hong, Yicong and Zhang, Kai and Gu, Jiuxiang and Bi, Sai and Zhou, Yang and Liu, Difan and Liu, Feng and Sunkavalli, Kalyan and Bui, Trung and Tan, Hao},
  journal={arXiv preprint arXiv:2311.04400},
  year={2023}
}

@inproceedings{chung2023promptpaint,
  title={Promptpaint: Steering text-to-image generation through paint medium-like interactions},
  author={Chung, John Joon Young and Adar, Eytan},
  booktitle={Proceedings of the 36th Annual ACM Symposium on User Interface Software and Technology},
  pages={1--17},
  year={2023}
}

@inproceedings{brade2023promptify,
  title={Promptify: Text-to-image generation through interactive prompt exploration with large language models},
  author={Brade, Stephen and Wang, Bryan and Sousa, Mauricio and Oore, Sageev and Grossman, Tovi},
  booktitle={Proceedings of the 36th Annual ACM Symposium on User Interface Software and Technology},
  pages={1--14},
  year={2023}
}

@inproceedings{zhu2020curveboards,
  title={CurveBoards: Integrating breadboards into physical objects to prototype function in the context of form},
  author={Zhu, Junyi and Blumberg, Lotta-Gili and Zhu, Yunyi and Nisser, Martin and Carlson, Ethan Levi and Wen, Xin and Shum, Kevin and Quaye, Jessica Ayeley and Mueller, Stefanie},
  booktitle={Proceedings of the 2020 CHI Conference on Human Factors in Computing Systems},
  pages={1--13},
  year={2020}
}

@inproceedings{shen2024neural,
  title={Neural canvas: Supporting scenic design prototyping by integrating 3d sketching and generative AI},
  author={Shen, Yulin and Shen, Yifei and Cheng, Jiawen and Jiang, Chutian and Fan, Mingming and Wang, Zeyu},
  booktitle={Proceedings of the 2024 CHI Conference on Human Factors in Computing Systems},
  pages={1--18},
  year={2024}
}

@article{faruqi2026compos3d,
  title={Compos3D: Interactive Part-Based Composition for Creative Control in Generative 3D Models},
  author={Faruqi, Faraz and Liu, Sean J and Fitzmaurice, George and Matejka, Justin},
  journal={arXiv preprint arXiv:2607.12193},
  year={2026}
}

@article{faruqi2026mixr,
  title={MiXR: Harvesting and Recomposing Geometry from Real-World Objects for In-Situ 3D Design},
  author={Faruqi, Faraz and Tas, Demircan and Caetano, Arthur and Meniconi, Niccol{\`o} and Arslan, O{\u{g}}uz and Sra, Misha and Du, Ruofei and Mueller, Stefanie and Dogan, Mustafa Doga},
  journal={arXiv preprint arXiv:2605.09620},
  year={2026}
}

@inproceedings{faruqi2025mechstyle,
  title={MechStyle: Augmenting Generative AI with Mechanical Simulation to Create Stylized and Structurally Viable 3D Models},
  author={Faruqi, Faraz and Abdel-Rahman, Amira and Tejedor, Leandra and Nisser, Martin and Li, Jiaji and Phadnis, Vrushank and Jampani, Varun and Gershenfeld, Neil and Hofmann, Megan and Mueller, Stefanie},
  booktitle={Proceedings of the ACM Symposium on Computational Fabrication},
  pages={1--15},
  year={2025}
}

@inproceedings{zhang2025a11yshape,
  title={A11yShape: AI-Assisted 3-D Modeling for Blind and Low-Vision Programmers},
  author={Zhang, Zhuohao and Li, Haichang and Yu, Chun Meng and Faruqi, Faraz and Xie, Junan and Kim, Gene SH and Fan, Mingming and Forbes, Angus and Wobbrock, Jacob O and Guo, Anhong and others},
  booktitle={Proceedings of the 27th International ACM SIGACCESS Conference on Computers and Accessibility},
  pages={1--20},
  year={2025}
}

@article{faruqi2024shaping,
  title={Shaping realities: Enhancing 3D generative AI with fabrication constraints},
  author={Faruqi, Faraz and Tian, Yingtao and Phadnis, Vrushank and Jampani, Varun and Mueller, Stefanie},
  journal={arXiv preprint arXiv:2404.10142},
  year={2024}
}

@inproceedings{chen2026sam,
  title={Sam 3d: 3dfy anything in images},
  author={Chen, Xingyu and Chu, Fu-Jen and Gleize, Pierre and Liang, Kevin J and Sax, Alexander and Tang, Hao and Wang, Weiyao and Guo, Michelle and Hardin, Thibaut and Li, Xiang and others},
  booktitle={Proceedings of the IEEE/CVF Conference on Computer Vision and Pattern Recognition},
  pages={7220--7232},
  year={2026}
}

@inproceedings{xiang2026native,
  title={Native and compact structured latents for 3d generation},
  author={Xiang, Jianfeng and Chen, Xiaoxue and Xu, Sicheng and Wang, Ruicheng and Lv, Zelong and Deng, Yu and Zhu, Hongyuan and Dong, Yue and Zhao, Hao and Yuan, Nicholas Jing and others},
  booktitle={Proceedings of the IEEE/CVF Conference on Computer Vision and Pattern Recognition},
  pages={14419--14429},
  year={2026}
}

@inproceedings{klokov2020discrete,
  title={Discrete point flow networks for efficient point cloud generation},
  author={Klokov, Roman and Boyer, Edmond and Verbeek, Jakob},
  booktitle={European Conference on Computer Vision},
  pages={694--710},
  year={2020},
  organization={Springer}
}

@inproceedings{zhang2023adding,
  title={Adding conditional control to text-to-image diffusion models},
  author={Zhang, Lvmin and Rao, Anyi and Agrawala, Maneesh},
  booktitle={Proceedings of the IEEE/CVF international conference on computer vision},
  pages={3836--3847},
  year={2023}
}

@inproceedings{dreameditor2023,
  title={Dreameditor: Text-driven 3d scene editing with neural fields},
  author={Zhuang, Jingyu and Wang, Chen and Lin, Liang and Liu, Lingjie and Li, Guanbin},
  booktitle={SIGGRAPH Asia 2023 conference papers},
  pages={1--10},
  year={2023}
}

@inproceedings{stemasov2024param,
  title={PARam: leveraging parametric design in extended reality to support the personalization of artifacts for personal fabrication},
  author={Stemasov, Evgeny and Demharter, Simon and R{\"a}dler, Max and Gugenheimer, Jan and Rukzio, Enrico},
  booktitle={Proceedings of the 2024 CHI Conference on Human Factors in Computing Systems},
  pages={1--22},
  year={2024}
}

@article{lu2024advances,
  title={Advances in text-guided 3D editing: a survey},
  author={Lu, Lihua and Li, Ruyang and Zhang, Xiaohui and Wei, Hui and Du, Guoguang and Wang, Binqiang},
  journal={Artificial Intelligence Review},
  volume={57},
  number={12},
  pages={321},
  year={2024},
  publisher={Springer}
}

@article{brooke1996sus,
  title={SUS-A quick and dirty usability scale},
  author={Brooke, John and others},
  journal={Usability evaluation in industry},
  volume={189},
  number={194},
  pages={4--7},
  year={1996},
  publisher={London, England}
}

@article{gpt4techreport,
  title={Gpt-4 technical report},
  author={Achiam, Josh and Adler, Steven and Agarwal, Sandhini and Ahmad, Lama and Akkaya, Ilge and Aleman, Florencia Leoni and Almeida, Diogo and Altenschmidt, Janko and Altman, Sam and Anadkat, Shyamal and others},
  journal={arXiv preprint arXiv:2303.08774},
  year={2023}
}

@misc{meshy2026,
  author = {{Meshy AI}},
  title = {Meshy: 3D AI Toolbox for Generative Mesh and Texture Creation},
  year = {2026},
  url = {https://www.meshy.ai/},
  note = {Version 6.0, Accessed: 2026-03-29}
}

@article{zhang2024clay,
  title={Clay: A controllable large-scale generative model for creating high-quality 3d assets},
  author={Zhang, Longwen and Wang, Ziyu and Zhang, Qixuan and Qiu, Qiwei and Pang, Anqi and Jiang, Haoran and Yang, Wei and Xu, Lan and Yu, Jingyi},
  journal={ACM Transactions on Graphics (TOG)},
  volume={43},
  number={4},
  pages={1--20},
  year={2024},
  publisher={ACM New York, NY, USA}
}

@article{zhao2025hunyuan3d,
  title={Hunyuan3d 2.0: Scaling diffusion models for high resolution textured 3d assets generation},
  author={Zhao, Zibo and Lai, Zeqiang and Lin, Qingxiang and Zhao, Yunfei and Liu, Haolin and Yang, Shuhui and Feng, Yifei and Yang, Mingxin and Zhang, Sheng and Yang, Xianghui and others},
  journal={arXiv preprint arXiv:2501.12202},
  year={2025}
}

@article{tipeditor2024,
  title={Tip-editor: An accurate 3d editor following both text-prompts and image-prompts},
  author={Zhuang, Jingyu and Kang, Di and Cao, Yan-Pei and Li, Guanbin and Lin, Liang and Shan, Ying},
  journal={ACM Transactions on Graphics (ToG)},
  volume={43},
  number={4},
  pages={1--12},
  year={2024},
  publisher={ACM New York, NY, USA}
}

@inproceedings{zamfirescu2023johnny,
  title={Why Johnny can’t prompt: how non-AI experts try (and fail) to design LLM prompts},
  author={Zamfirescu-Pereira, J Diego and Wong, Richmond Y and Hartmann, Bjoern and Yang, Qian},
  booktitle={Proceedings of the 2023 CHI conference on human factors in computing systems},
  pages={1--21},
  year={2023}
}

@article{subramonyam2023bridging,
  title={Bridging the Gulf of envisioning: Cognitive design challenges in LLM interfaces},
  author={Subramonyam, Hariharan and Pea, Roy and Pondoc, Christopher Lawrence and Agrawala, Maneesh and Seifert, Colleen},
  journal={arXiv preprint arXiv:2309.14459},
  year={2023}
}

@misc{trimesh2019,
  author       = {Michael Dawson-Haggerty},
  title        = {Trimesh},
  year         = {2019},
  howpublished = {\url{https://github.com/mikedh/trimesh}},
  note         = {Computer software},
}

@inproceedings{liu20233dall,
  title={3DALL-E: Integrating text-to-image AI in 3D design workflows},
  author={Liu, Vivian and Vermeulen, Jo and Fitzmaurice, George and Matejka, Justin},
  booktitle={Proceedings of the 2023 ACM designing interactive systems conference},
  pages={1955--1977},
  year={2023}
}

@inproceedings{chou2023diffusion,
  title={Diffusion-sdf: Conditional generative modeling of signed distance functions},
  author={Chou, Gene and Bahat, Yuval and Heide, Felix},
  booktitle={Proceedings of the IEEE/CVF international conference on computer vision},
  pages={2262--2272},
  year={2023}
}

@article{wu2019sagnet,
  title={Sagnet: Structure-aware generative network for 3d-shape modeling},
  author={Wu, Zhijie and Wang, Xiang and Lin, Di and Lischinski, Dani and Cohen-Or, Daniel and Huang, Hui},
  journal={ACM Transactions on Graphics (TOG)},
  volume={38},
  number={4},
  pages={1--14},
  year={2019},
  publisher={ACM New York, NY, USA}
}

@inproceedings{yang2019pointflow,
  title={Pointflow: 3d point cloud generation with continuous normalizing flows},
  author={Yang, Guandao and Huang, Xun and Hao, Zekun and Liu, Ming-Yu and Belongie, Serge and Hariharan, Bharath},
  booktitle={Proceedings of the IEEE/CVF international conference on computer vision},
  pages={4541--4550},
  year={2019}
}

@inproceedings{mittal2022autosdf,
  title={Autosdf: Shape priors for 3d completion, reconstruction and generation},
  author={Mittal, Paritosh and Cheng, Yen-Chi and Singh, Maneesh and Tulsiani, Shubham},
  booktitle={Proceedings of the IEEE/CVF conference on computer vision and pattern recognition},
  pages={306--315},
  year={2022}
}

@inproceedings{cai2020learning,
  title={Learning gradient fields for shape generation},
  author={Cai, Ruojin and Yang, Guandao and Averbuch-Elor, Hadar and Hao, Zekun and Belongie, Serge and Snavely, Noah and Hariharan, Bharath},
  booktitle={Computer Vision--ECCV 2020: 16th European Conference, Glasgow, UK, August 23--28, 2020, Proceedings, Part III 16},
  pages={364--381},
  year={2020},
  organization={Springer}
}

@inproceedings{wu2022ai,
  title={Ai chains: Transparent and controllable human-ai interaction by chaining large language model prompts},
  author={Wu, Tongshuang and Terry, Michael and Cai, Carrie Jun},
  booktitle={Proceedings of the 2022 CHI conference on human factors in computing systems},
  pages={1--22},
  year={2022}
}

@inproceedings{kim2024evallm,
  title={Evallm: Interactive evaluation of large language model prompts on user-defined criteria},
  author={Kim, Tae Soo and Lee, Yoonjoo and Shin, Jamin and Kim, Young-Ho and Kim, Juho},
  booktitle={Proceedings of the 2024 CHI Conference on Human Factors in Computing Systems},
  pages={1--21},
  year={2024}
}

@inproceedings{reza2024abscribe,
  title={ABScribe: Rapid Exploration \& Organization of Multiple Writing Variations in Human-AI Co-Writing Tasks using Large Language Models},
  author={Reza, Mohi and Laundry, Nathan M and Musabirov, Ilya and Dushniku, Peter and Yu, Zhi Yuan “Michael” and Mittal, Kashish and Grossman, Tovi and Liut, Michael and Kuzminykh, Anastasia and Williams, Joseph Jay},
  booktitle={Proceedings of the 2024 CHI Conference on Human Factors in Computing Systems},
  pages={1--18},
  year={2024}
}

@inproceedings{wang2024promptcharm,
  title={Promptcharm: Text-to-image generation through multi-modal prompting and refinement},
  author={Wang, Zhijie and Huang, Yuheng and Song, Da and Ma, Lei and Zhang, Tianyi},
  booktitle={Proceedings of the 2024 CHI Conference on Human Factors in Computing Systems},
  pages={1--21},
  year={2024}
}

@article{boss2024sf3d,
  title={Sf3d: Stable fast 3d mesh reconstruction with uv-unwrapping and illumination disentanglement},
  author={Boss, Mark and Huang, Zixuan and Vasishta, Aaryaman and Jampani, Varun},
  journal={arXiv preprint arXiv:2408.00653},
  year={2024}
}

@inproceedings{tang2024lgm,
  title={Lgm: Large multi-view gaussian model for high-resolution 3d content creation},
  author={Tang, Jiaxiang and Chen, Zhaoxi and Chen, Xiaokang and Wang, Tengfei and Zeng, Gang and Liu, Ziwei},
  booktitle={European Conference on Computer Vision},
  pages={1--18},
  year={2024},
  organization={Springer}
}

@article{faruqi2025tactstyle,
  title={TactStyle: Generating Tactile Textures with Generative AI for Digital Fabrication},
  author={Faruqi, Faraz and Perroni-Scharf, Maxine and Walia, Jaskaran Singh and Zhu, Yunyi and Feng, Shuyue and Degraen, Donald and Mueller, Stefanie},
  journal={arXiv preprint arXiv:2503.02007},
  year={2025}
}

@article{xu2024instantmesh,
  title={Instantmesh: Efficient 3d mesh generation from a single image with sparse-view large reconstruction models},
  author={Xu, Jiale and Cheng, Weihao and Gao, Yiming and Wang, Xintao and Gao, Shenghua and Shan, Ying},
  journal={arXiv preprint arXiv:2404.07191},
  year={2024}
}

@inproceedings{AI-music,
  title={Novice-AI music co-creation via AI-steering tools for deep generative models},
  author={Louie, Ryan and Coenen, Andy and Huang, Cheng Zhi and Terry, Michael and Cai, Carrie J},
  booktitle={Proceedings of the 2020 CHI conference on human factors in computing systems},
  pages={1--13},
  year={2020}
}

@article{jun2023shap_e,
  title={Shap-e: Generating conditional 3d implicit functions},
  author={Jun, Heewoo and Nichol, Alex},
  journal={arXiv preprint arXiv:2305.02463},
  year={2023}
}

@inproceedings{hudson2016understanding,
author = {Hudson, Nathaniel and Alcock, Celena and Chilana, Parmit K.},
title = {Understanding Newcomers to 3D Printing: Motivations, Workflows, and Barriers of Casual Makers},
year = {2016},
isbn = {9781450333627},
publisher = {Association for Computing Machinery},
address = {New York, NY, USA},
url = {https://doi.org/10.1145/2858036.2858266},
doi = {10.1145/2858036.2858266},
booktitle = {Proceedings of the 2016 CHI Conference on Human Factors in Computing Systems},
pages = {384–396},
numpages = {13},
location = {San Jose, California, USA},
series = {CHI '16}
}

@inproceedings{buehler2015sharing,
author = {Buehler, Erin and Branham, Stacy and Ali, Abdullah and Chang, Jeremy J. and Hofmann, Megan Kelly and Hurst, Amy and Kane, Shaun K.},
title = {Sharing is Caring: Assistive Technology Designs on Thingiverse},
year = {2015},
isbn = {9781450331456},
publisher = {Association for Computing Machinery},
address = {New York, NY, USA},
url = {https://doi.org/10.1145/2702123.2702525},
doi = {10.1145/2702123.2702525},
booktitle = {Proceedings of the 33rd Annual ACM Conference on Human Factors in Computing Systems},
pages = {525–534},
numpages = {10},
location = {Seoul, Republic of Korea},
series = {CHI '15}
}

@article{schmidt2013design,
  author={Schmidt, Ryan and Ratto, Matt},
  journal={IEEE Computer Graphics and Applications}, 
  title={Design-to-Fabricate: Maker Hardware Requires Maker Software}, 
  year={2013},
  volume={33},
  number={6},
  pages={26-34},
  doi={10.1109/MCG.2013.90}}

@article{schulz2014design,
author = {Schulz, Adriana and Shamir, Ariel and Levin, David I. W. and Sitthi-amorn, Pitchaya and Matusik, Wojciech},
title = {Design and Fabrication by Example},
year = {2014},
issue_date = {July 2014},
publisher = {Association for Computing Machinery},
address = {New York, NY, USA},
volume = {33},
number = {4},
issn = {0730-0301},
url = {https://doi.org/10.1145/2601097.2601127},
doi = {10.1145/2601097.2601127},
journal = {ACM Trans. Graph.},
month = {jul},
articleno = {62},
numpages = {11}
}

@article{shugrina2015fab,
author = {Shugrina, Maria and Shamir, Ariel and Matusik, Wojciech},
title = {Fab Forms: Customizable Objects for Fabrication with Validity and Geometry Caching},
year = {2015},
issue_date = {August 2015},
publisher = {Association for Computing Machinery},
address = {New York, NY, USA},
volume = {34},
number = {4},
issn = {0730-0301},
url = {https://doi.org/10.1145/2766994},
doi = {10.1145/2766994},
journal = {ACM Trans. Graph.},
month = {jul},
articleno = {100},
numpages = {12}
}

@inproceedings{berman2021howdiy,
author = {Berman, Alexander and Thakare, Ketan and Howell, Joshua and Quek, Francis and Kim, Jeeeun},
title = {HowDIY: Towards Meta-Design Tools to Support Anyone to 3D Print Anywhere},
year = {2021},
isbn = {9781450380171},
publisher = {Association for Computing Machinery},
address = {New York, NY, USA},
url = {https://doi.org/10.1145/3397481.3450638},
doi = {10.1145/3397481.3450638},
booktitle = {26th International Conference on Intelligent User Interfaces},
pages = {491–503},
numpages = {13},
location = {College Station, TX, USA},
series = {IUI '21}
}

@inproceedings{norouzi2021making,
author = {Norouzi, Behnaz and Kinnula, Marianne and Iivari, Netta},
title = {Making Sense of 3D Modelling and 3D Printing Activities of Young People: A Nexus Analytic Inquiry},
year = {2021},
isbn = {9781450380966},
publisher = {Association for Computing Machinery},
address = {New York, NY, USA},
url = {https://doi.org/10.1145/3411764.3445139},
doi = {10.1145/3411764.3445139},
booktitle = {Proceedings of the 2021 CHI Conference on Human Factors in Computing Systems},
articleno = {481},
numpages = {16},
location = {Yokohama, Japan},
series = {CHI '21}
}

@inproceedings{veuskens2020coda,
title={{CODA}: A Design Assistant to Facilitate Specifying Constraints and Parametric Behavior in {CAD} Models},
author={Tom Veuskens and Florian Heller and Raf Ramakers},
booktitle={Graphics Interface 2021},
year={2021},
url={https://openreview.net/forum?id=1dLDPJeafRZ}
}

@article{prevost2013make,
author = {Pr\'{e}vost, Romain and Whiting, Emily and Lefebvre, Sylvain and Sorkine-Hornung, Olga},
title = {Make It Stand: Balancing Shapes for 3D Fabrication},
year = {2013},
issue_date = {July 2013},
publisher = {Association for Computing Machinery},
address = {New York, NY, USA},
volume = {32},
number = {4},
issn = {0730-0301},
url = {https://doi.org/10.1145/2461912.2461957},
doi = {10.1145/2461912.2461957},
journal = {ACM Trans. Graph.},
month = {jul},
articleno = {81},
numpages = {10}
}

@article{chowdhery2023palm,
  title={Palm: Scaling language modeling with pathways},
  author={Chowdhery, Aakanksha and Narang, Sharan and Devlin, Jacob and Bosma, Maarten and Mishra, Gaurav and Roberts, Adam and Barham, Paul and Chung, Hyung Won and Sutton, Charles and Gehrmann, Sebastian and others},
  journal={Journal of Machine Learning Research},
  volume={24},
  number={240},
  pages={1--113},
  year={2023}
}

@article{bacher2014spin,
author = {B\"{a}cher, Moritz and Whiting, Emily and Bickel, Bernd and Sorkine-Hornung, Olga},
title = {Spin-It: Optimizing Moment of Inertia for Spinnable Objects},
year = {2014},
issue_date = {July 2014},
publisher = {Association for Computing Machinery},
address = {New York, NY, USA},
volume = {33},
number = {4},
issn = {0730-0301},
url = {https://doi.org/10.1145/2601097.2601157},
doi = {10.1145/2601097.2601157},
journal = {ACM Trans. Graph.},
month = {jul},
articleno = {96},
numpages = {10}
}

@article{zhao2016make,
title = {Make it swing: Fabricating personalized roly-poly toys},
journal = {Computer Aided Geometric Design},
volume = {43},
pages = {226-236},
year = {2016},
note = {Geometric Modeling and Processing 2016},
issn = {0167-8396},
doi = {https://doi.org/10.1016/j.cagd.2016.02.001},
url = {https://www.sciencedirect.com/science/article/pii/S0167839616300024},
author = {Haiming Zhao and Chengkuan Hong and Juncong Lin and Xiaogang Jin and Weiwei Xu}
}

@inproceedings{prevost2016balancing,
author = {Pr\'{e}vost, Romain and B\"{a}cher, Moritz and Jarosz, Wojciech and Sorkine-Hornung, Olga},
title = {Balancing 3D Models with Movable Masses},
year = {2016},
isbn = {9783038680253},
publisher = {Eurographics Association},
address = {Goslar, DEU},
booktitle = {Proceedings of the Conference on Vision, Modeling and Visualization},
pages = {9–16},
numpages = {8},
location = {Bayreuth, Germany},
series = {VMV '16}
}

@article{zhang2017functionality,
author = {Zhang, Ran and Auzinger, Thomas and Ceylan, Duygu and Li, Wilmot and Bickel, Bernd},
title = {Functionality-Aware Retargeting of Mechanisms to 3D Shapes},
year = {2017},
issue_date = {August 2017},
publisher = {Association for Computing Machinery},
address = {New York, NY, USA},
volume = {36},
number = {4},
issn = {0730-0301},
url = {https://doi.org/10.1145/3072959.3073710},
doi = {10.1145/3072959.3073710},
journal = {ACM Trans. Graph.},
month = {jul},
articleno = {81},
numpages = {13}
}

@inproceedings{oehlberg2015patterns,
author = {Oehlberg, Lora and Willett, Wesley and Mackay, Wendy E.},
title = {Patterns of Physical Design Remixing in Online Maker Communities},
year = {2015},
isbn = {9781450331456},
publisher = {Association for Computing Machinery},
address = {New York, NY, USA},
url = {https://doi.org/10.1145/2702123.2702175},
doi = {10.1145/2702123.2702175},
booktitle = {Proceedings of the 33rd Annual ACM Conference on Human Factors in Computing Systems},
pages = {639–648},
numpages = {10},
location = {Seoul, Republic of Korea},
series = {CHI '15}
}

@inproceedings{Kuznetsov_2010_expertamateur,
 author = {Kuznetsov, Stacey and Paulos, Eric},
 title = {Rise of the Expert Amateur: DIY Projects, Communities, and Cultures},
 booktitle = {Proceedings of the 6th Nordic Conference on Human-Computer Interaction: Extending Boundaries},
 series = {NordiCHI '10},
 year = {2010},
 isbn = {978-1-60558-934-3},
  
 pages = {295--304},
 numpages = {10},
 url = {http://doi.acm.org/10.1145/1868914.1868950},
 doi = {10.1145/1868914.1868950},
 acmid = {1868950},
 publisher = {ACM},
}

@Manual{blenderUI,
   title = {Blender - a 3D modelling and rendering package},
   author = {Blender Online Community},
   organization = {Blender Foundation},
   address = {Stichting Blender Foundation, Amsterdam},
   year = {2018},
   url = {http://www.blender.org},
 }

@article{Lakshmi_poc,
author = {Lakshmi, Udaya and Hofmann, Megan and Valencia, Stephanie and Wilcox, Lauren and Mankoff, Jennifer and Arriaga, Rosa I.},
title = {"Point-of-Care Manufacturing": Maker Perspectives on Digital Fabrication in Medical Practice},
year = {2019},
issue_date = {November 2019},
publisher = {Association for Computing Machinery},
address = {New York, NY, USA},
volume = {3},
number = {CSCW},
url = {https://doi.org/10.1145/3359193},
doi = {10.1145/3359193},
journal = {Proc. ACM Hum.-Comput. Interact.},
month = {nov},
articleno = {91},
numpages = {23}
}

@article{Shinohara_social,
author = {Shinohara, Kristen and Bennett, Cynthia L. and Pratt, Wanda and Wobbrock, Jacob O.},
title = {Tenets for Social Accessibility: Towards Humanizing Disabled People in Design},
year = {2018},
issue_date = {March 2018},
publisher = {Association for Computing Machinery},
address = {New York, NY, USA},
volume = {11},
number = {1},
issn = {1936-7228},
url = {https://doi.org/10.1145/3178855},
doi = {10.1145/3178855},
journal = {ACM Trans. Access. Comput.},
month = {mar},
articleno = {6},
numpages = {31}
}

@article{koyama2015autoconnect,
author = {Koyama, Yuki and Sueda, Shinjiro and Steinhardt, Emma and Igarashi, Takeo and Shamir, Ariel and Matusik, Wojciech},
title = {AutoConnect: Computational Design of 3D-Printable Connectors},
year = {2015},
issue_date = {November 2015},
publisher = {Association for Computing Machinery},
address = {New York, NY, USA},
volume = {34},
number = {6},
issn = {0730-0301},
url = {https://doi.org/10.1145/2816795.2818060},
doi = {10.1145/2816795.2818060},
journal = {ACM Trans. Graph.},
month = {nov},
articleno = {231},
numpages = {11}
}

@inproceedings{roumen2018grafter,
author = {Roumen, Thijs Jan and M\"{u}ller, Willi and Baudisch, Patrick},
title = {Grafter: Remixing 3D-Printed Machines},
year = {2018},
isbn = {9781450356206},
publisher = {Association for Computing Machinery},
address = {New York, NY, USA},
url = {https://doi.org/10.1145/3173574.3173637},
doi = {10.1145/3173574.3173637},
booktitle = {Proceedings of the 2018 CHI Conference on Human Factors in Computing Systems},
pages = {1–12},
numpages = {12},
location = {Montreal QC, Canada},
series = {CHI '18}
}

@incollection{schmidt2010meshmixer,
author = {Schmidt, Ryan and Singh, Karan},
title = {Meshmixer: An Interface for Rapid Mesh Composition},
year = {2010},
isbn = {9781450303941},
publisher = {Association for Computing Machinery},
address = {New York, NY, USA},
url = {https://doi.org/10.1145/1837026.1837034},
doi = {10.1145/1837026.1837034},
booktitle = {ACM SIGGRAPH 2010 Talks},
articleno = {6},
numpages = {1},
location = {Los Angeles, California},
series = {SIGGRAPH '10}
}

@software{pymeshlab,
  author       = {Alessandro Muntoni and Paolo Cignoni},
  title        = {{PyMeshLab}},
  month        = jan,
  year         = 2021,
  publisher    = {Zenodo},
  doi          = {10.5281/zenodo.4438750}
}

@article{faruqi2021slicehub,
  title={SliceHub: Augmenting Shared 3D Model Repositories with Slicing Results for 3D Printing},
  author={Faruqi, Faraz and Friedman, Kenneth and Cheng, Leon and Wessely, Michael and Subramanian, Sriram and Mueller, Stefanie},
  journal={arXiv preprint arXiv:2109.14722},
  year={2021}
}

@inproceedings{faruqi2023style2fab,
  title={Style2Fab: Functionality-Aware Segmentation for Fabricating Personalized 3D Models with Generative AI},
  author={Faruqi, Faraz and Katary, Ahmed and Hasic, Tarik and Abdel-Rahman, Amira and Rahman, Nayeemur and Tejedor, Leandra and Leake, Mackenzie and Hofmann, Megan and Mueller, Stefanie},
  booktitle={Proceedings of the 36th Annual ACM Symposium on User Interface Software and Technology},
  pages={1--13},
  year={2023}
}

@inproceedings{gao2022get3d,
title={GET3D: A Generative Model of High Quality 3D Textured Shapes Learned from Images},
author={Jun Gao and Tianchang Shen and Zian Wang and Wenzheng Chen and Kangxue Yin
and Daiqing Li and Or Litany and Zan Gojcic and Sanja Fidler},
booktitle={Advances In Neural Information Processing Systems},
year={2022}
}

@inproceedings{rombach2022high,
  title={High-resolution image synthesis with latent diffusion models},
  author={Rombach, Robin and Blattmann, Andreas and Lorenz, Dominik and Esser, Patrick and Ommer, Bj{\"o}rn},
  booktitle={Proceedings of the IEEE/CVF conference on computer vision and pattern recognition},
  pages={10684--10695},
  year={2022}
}

@misc{xiang2024structured3dlatentsscalable,
      title={Structured 3D Latents for Scalable and Versatile 3D Generation}, 
      author={Jianfeng Xiang and Zelong Lv and Sicheng Xu and Yu Deng and Ruicheng Wang and Bowen Zhang and Dong Chen and Xin Tong and Jiaolong Yang},
      year={2024},
      eprint={2412.01506},
      archivePrefix={arXiv},
      primaryClass={cs.CV},
      url={https://arxiv.org/abs/2412.01506}, 
}

@misc{huang2025spar3dstablepointawarereconstruction,
      title={SPAR3D: Stable Point-Aware Reconstruction of 3D Objects from Single Images}, 
      author={Zixuan Huang and Mark Boss and Aaryaman Vasishta and James M. Rehg and Varun Jampani},
      year={2025},
      eprint={2501.04689},
      archivePrefix={arXiv},
      primaryClass={cs.CV},
      url={https://arxiv.org/abs/2501.04689}, 
}

\newpage


\end{document}